\pdfoutput=1
\documentclass[sigconf,nonacm]{acmart}

\usepackage{bm}
\usepackage{microtype}
\usepackage{graphicx}
\usepackage{booktabs}
\usepackage{amsmath}
\usepackage{mathtools}
\usepackage{amsthm}
\usepackage{multirow}
\usepackage{enumitem}
\usepackage{xspace}
\usepackage{wrapfig}
\usepackage{tcolorbox}
\usepackage{colortbl}
\usepackage{subcaption}

\theoremstyle{plain}
\newtheorem{theorem}{Theorem}[section]
\newtheorem{proposition}[theorem]{Proposition}

\newtheorem{corollary}[theorem]{Corollary}
\theoremstyle{definition}
\newtheorem{definition}[theorem]{Definition}

\theoremstyle{remark}

\newcommand{\gG}{\mathcal{G}}
\newcommand{\gV}{\mathcal{V}}
\newcommand{\gE}{\mathcal{E}}
\newcommand{\mA}{\mathbf{A}}
\newcommand{\mB}{\mathbf{B}}
\newcommand{\mD}{\mathbf{D}}
\newcommand{\mX}{\mathbf{X}}
\newcommand{\mW}{\mathbf{W}}
\newcommand{\mU}{\mathbf{U}}
\newcommand{\mV}{\mathbf{V}}
\newcommand{\vx}{\mathbf{x}}
\newcommand{\vh}{\mathbf{h}}
\newcommand{\vy}{\mathbf{y}}
\newcommand{\vz}{\mathbf{z}}
\newcommand{\vs}{\mathbf{s}}

\newcommand{\va}{\mathbf{a}}

\newcommand{\method}{\textsc{SCOPE}\xspace}

\newcommand{\methoddiag}{\method-diag}

\begin{document}
\title{Learning Graph-Conditioned Predictors in Weight Space}

\author{Zhewei Chen}
\affiliation{%
  \institution{Australian National University}
  \country{Australia}
}
\email{zhewei.chen@anu.edu.au}

\author{Hao Zhu}
\affiliation{%
  \institution{CSIRO Technology}
  \country{Australia}
}
\email{allenhaozhu@gmail.com}

\author{Ahad N. Zehmakan}
\affiliation{%
  \institution{Australian National University}
  \country{Australia}
}
\email{ahadn.zehmakan@anu.edu.au}

\renewcommand{\shortauthors}{Chen et al.}

\begin{abstract}
The same node attributes can imply different labels in different structural contexts. Yet fast
graphless models typically treat topology as additional evidence supplied to a globally shared
predictor. This assumes that structure augments the available information, rather than determining
how node attributes should be interpreted.
We introduce \method (Structure-COnditioned PrEdictor), in which
a small generator maps each node's cached structural fingerprint to coefficients over a
shared low-rank basis, producing a context-specific update to a global weight matrix. The base
predictor, basis, and generator remain shared across all nodes, while the resulting transformation
adapts how attributes are processed in each relational environment. Once fingerprints are cached,
inference requires only a forward pass and no online message passing.
This design combines context-specific specialization with global parameter sharing, compact
low-rank adaptation, and message-passing-free inference.
On two large OGB node-classification graphs, conditioning the weights with raw positional
signals yields clear gains over supplying the same signals through input concatenation. Further
results under a structure-transductive held-out-node protocol and in teacher-free graph
classification show that the mechanism extends beyond a particular distillation setting. These
findings establish graph structure not only as predictive evidence, but as context that determines
the predictor used to interpret that evidence.
\end{abstract}


\maketitle

\section{Introduction}
\label{sec:intro}

Two nodes in a citation network can share nearly identical textual features (keywords, venue,
co‑authors), yet their structural roles can make those features mean entirely different things.  A
bridging paper that connects several communities and a leaf that cites only one cluster might
require opposite predictions from the same input pattern.  This is a familiar challenge in graph learning: the optimal mapping from features to labels often depends on local topology, as work on heterophilous graphs has shown when connected nodes carry dissimilar labels and one globally shared aggregation rule no longer suffices~\cite{zhu2020beyond}.  Yet we
rarely deploy a separate model for every node; we train a single shared predictor that pools
supervision and generalises to unseen instances.  The problem is that one fixed function must cover very different relational environments.

Message-passing GNNs make computation inherently structure-aware, and graph mixture-of-experts
models specialise it further through node-dependent routing~\cite{wang2023gmoe}. In
latency-sensitive settings, however, deployment often favors graphless MLPs that avoid online
neighborhood aggregation~\cite{zhang2022graphless}. Distillation, positional encodings, and
precomputed propagation features can supply these models with structural information
~\cite{yang2023vqgraph,wu2023krd,tian2023nosmog,chen2021gamlp}, but they typically do so by changing
the model's inputs or supervision while retaining a fixed set of operators. This distinction matters
when topology changes how attributes relate to the target: the same feature direction may need to be
amplified for a hub but suppressed for a peripheral node. A fixed predictor must encode these
competing mappings within one shared transformation, potentially weakening the role-specific signal
that the structural information was intended to provide.

\method\ takes a different route: structure changes the predictor's effective parameters during each
forward pass. A small generator reads a cached structural fingerprint, such as degree, PageRank,
or positional coordinates, and outputs coefficients over shared rank-one basis matrices. Their
weighted combination forms a compact low-rank update to a global base matrix, so different
structural contexts induce different linear operators on the same node attributes. This produces a
family of context-specific predictors without learning or storing an independent model for every
node: the base matrix, basis, and generator remain globally shared and are trained end to end. Once
the fingerprints are cached, prediction remains a standard forward computation with no online
neighborhood aggregation or graph-operator evaluation.

The low-rank construction makes this adaptation compact and structured. Rather than generating a
full weight matrix for every node, the generator only selects a mixture of shared adjustment
patterns. The basis captures recurring modes of parameter change, while the generator determines
which modes are relevant to the current structural context and how strongly they should be applied.
This factorization controls parameter growth and allows related contexts to reuse common
transformations instead of learning separate predictors. A baseline-preserving initialization sets
the generated update to zero, so the initial model is functionally identical to the plain shared MLP.
Training can then introduce context dependence gradually while retaining the shared predictor as its
reference point.

Our work evaluates this conditioning principle through carefully controlled experiments that
disentangle the effects of extra information, extra capacity, and the specific role that structure
plays.  The main findings are that (1) meaningful structural conditioning, not weight generation
alone, drives the improvement; (2)
the separation is largest on the two OGB graphs, which is consistent with---but does not by itself isolate---an effect of broader structural-context diversity;
(3) the generated weights form a coherent family organised with respect to structural statistics and task labels; and (4) the mechanism generalises beyond distillation, enabling teacher‑free graph classification.
\textbf{Our contributions are listed as follows:}

\begin{itemize}[leftmargin=*]

\item We formulate learning across different relational environments
as a shared-adaptation problem: graph structure is used not only
as extra predictive evidence, but also to determine how a shared
predictor processes each instance.

\item \method maps each cached structural fingerprint to a compact
low-rank update of a shared base model, producing instance-conditioned
effective weights while preserving parameter sharing and message-passing-free inference once fingerprints are cached.

\item We characterize the proposed parameterization theoretically. We show how it preserves graph
symmetry, supports operator changes beyond row-wise modulation, and approximates
context-dependent predictors through a shared low-rank basis.

\item Experiments across node and graph classification demonstrate the effectiveness of
structure-conditioned weight adaptation, with the strongest gains on the two OGB graphs. The
learned predictors organize by structural context, and the mechanism extends to teacher-free graph
classification.

\end{itemize}

\section{Related Work}

\label{sec:related}

\noindent\textbf{Weight-space learning and conditional parameter generation.}
Weight-space models process trained parameters while respecting neuron-permutation symmetries
\cite{navon2023dwsnets,zhou2023nfn,lim2024gmn}, with expressivity studied in
\cite{dayan2026expressive,sverdlov2024irreps}. A complementary line generates parameters:
HyperNetworks map conditioning codes to target-network weights \cite{ha2017hypernetworks}, dynamic filters generate input-dependent operators \cite{debrabandere2016dynamic}, and CondConv mixes shared kernels using input-dependent coefficients \cite{yang2019condconv}, the convolutional form of a mixture-of-experts layer in which a learned gate combines a shared bank of experts \cite{shazeer2017moe}. Graph HyperNetworks instead condition on the target architecture
\cite{zhang2019ghn}, while generative weight models learn distributions over parameters
\cite{peebles2022gpt,wang2024pdiff}. Most directly, Mapping Networks map a compact free latent to
target weights and provide a loss-matching existence result \cite{sen2026mapping}. FiLM
\cite{perez2018film,dumoulin2018featurewise} conditions a network by applying feature-wise scales
and shifts to its activations, while LoRA~\cite{hu2022lora} represents parameter adaptation through
low-rank weight updates. In contrast to row-wise FiLM modulation, \method\ uses graph context to
generate low-rank changes to the feature-processing operator itself, enabling operator directions
beyond rescaling the rows of a fixed base transformation. Feature-wise modulation has also been instantiated
directly on graphs: GNN-FiLM~\cite{brockschmidt2020gnnfilm} generates per-edge-type scale and shift
parameters and applies them to the incoming messages, so the modulation happens inside online
aggregation and therefore does not meet the message-passing-free inference constraint we impose here (Def.~\ref{def:graph_free}).
\method\ differs by using graph-derived relational
context to select compact weight updates within a shared learned basis, thereby learning a family of
context-specific predictors.

\noindent\textbf{Graph-to-MLP interfaces.}
Existing methods transfer graph information to MLP predictors through supervision, initialization,
or additional inputs. GLNN distills a message-passing teacher into a shared MLP
\cite{zhang2022graphless}; VQGraph, KRD, FF-G2M, and Cold Brew transfer richer representations,
relations, or discrete codes
\cite{yang2023vqgraph,wu2023krd,wu2023ffg2m,zheng2022cold}, while TINED injects teacher parameters
and distills layerwise behavior \cite{zhou2025tined}. Other methods expose structure directly through
DeepWalk positions, adjacency encoders, or precomputed propagation features
\cite{tian2023nosmog,chen2022samlp,chen2021gamlp,wu2019simplifying,frasca2020sign}.
Across these routes, structure changes the information, supervision, or initialization available to
a globally shared predictor. \method\ instead uses relational context to adapt the predictor's
effective weights, and applies the same mechanism with or without distillation.

\begin{figure}[t]
\centering
\includegraphics[width=\linewidth]{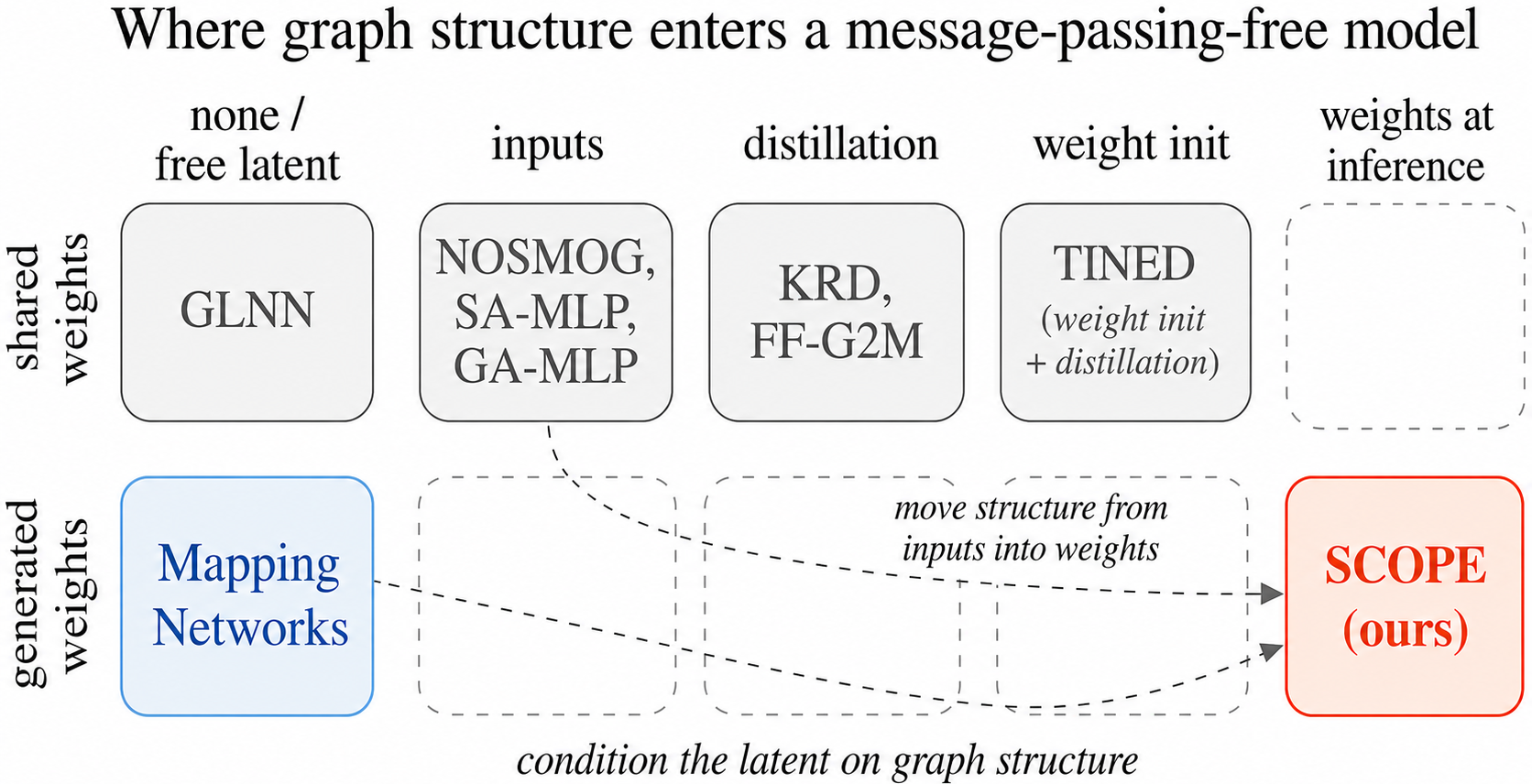}
\caption{Existing methods inject structure via inputs, supervision, or initialization, or generate weights from an unconditioned latent; \method instead maps cached structural fingerprints to context-dependent effective weights without online neighborhood aggregation.}

\label{fig:positioning}
\vspace{-1em}
\end{figure}

\section{Method}
\label{sec:method}

This section presents \method\ as a mechanism for balancing parameter sharing with context-specific specialization. Rather than learning an independent predictor for every node or relying on one fixed predictor across all relational environments, \method\ uses structural context to select a member of a shared low-dimensional family of effective predictors. We first formalize this distinction through three forms of message-passing-free structural conditioning (i.e., using only cached structural fingerprints at inference), then introduce an efficient low-rank weight-generation mechanism. We subsequently specify the structural fingerprints and training objectives, extend the construction to graph classification, and analyze its symmetry, representational advantages, and shared-basis approximation properties.
\subsection{Problem Setting: Relational Heterogeneity and Shared Adaptation}
\label{sec:prelim}

Let $\gG=(\gV,\gE)$ be a graph with node features $\mX\in\mathbb{R}^{|\gV|\times d}$, where
$\vx_v\in\mathbb{R}^d$ denotes the feature vector of node $v\in\gV$.
Nodes in the same graph may occupy heterogeneous relational environments, so the predictive relevance of a node’s attributes can depend on its structural context. Learning an independent predictor for every node would discard parameter sharing and would not generalize to unseen instances. We therefore seek a shared predictor that can nevertheless adapt its feature processing according to a compact structural fingerprint.

\begin{definition}[Message-passing-free inference]
\label{def:graph_free}
A predictor satisfies \emph{message-passing-free inference} if it performs no online neighborhood
aggregation or graph-operator computation during its neural forward pass.  At inference, it
receives the node feature vector $\vx_v$ and may additionally receive a structural
fingerprint (descriptor) $\vz_v$ computed during a separate graph-preprocessing stage.
\end{definition}

\noindent Under this deployment constraint, three natural designs arise according to whether structural context is omitted, supplied as additional predictive evidence, or used to adapt the predictor itself:
\begin{align}
\hat{\vy}_v &= f(\vx_v;\,\theta) \tag{M1}\label{eq:m1}\\[2pt]
\hat{\vy}_v &= f([\vx_v,\vz_v];\,\theta) \tag{M2}\label{eq:m2}\\[2pt]
\hat{\vy}_v &= f\!\left(\vx_v;\,\theta(\vz_v)\right) \tag{M3}\label{eq:m3}
\end{align}

\noindent\textbf{\eqref{eq:m1} Feature-only shared predictor.}
The simplest student applies one globally shared MLP to every node, with $\theta$ fixed across $\gV$.  It pools supervision efficiently across instances but receives no structural context at inference.  Consequently, two nodes with identical $\vx_v$ but different relational environments receive identical predictions, even when their attributes should be interpreted differently.  This is the basic deployed student of GLNN~\cite{zhang2022graphless}.

\noindent\textbf{\eqref{eq:m2} Structure as predictive evidence.}
Appending a cached structural fingerprint $\vz_v$ exposes the relational environment to the shared predictor.  An MLP receiving $[\vx_v,\vz_v]$ is conditionally expressive and can in principle represent different predictions across structural contexts.  However, it retains one globally parameterized operator: structural context can alter attribute processing only implicitly through the hidden activations, with no explicit factorization in which the context selects how the predictor should transform $\vx_v$.

\noindent\textbf{\eqref{eq:m3} Structure as adaptation context (family; ours is a learned-basis member).}
We instead use $\vz_v$ to select the effective operators applied to $\vx_v$.  This defines a shared family of context-specific predictors: all nodes reuse the same underlying parameters and adaptation mechanism, while different relational environments can induce different feature transformations through an explicit architectural pathway.  The challenge is to realize this family without generating or storing a full independent model for every node.  The following subsection develops an efficient low-rank construction.

\eqref{eq:m3} defines a family whose members differ in how $\Delta\mW(\vz)$ is parameterised
around a shared base $\mW_0$: \eqref{eq:m1}/\eqref{eq:m2} correspond to a zero or bias-only
update; the tied-basis member (\methoddiag) gives $\Delta\mW=\operatorname{diag}(\gamma(\vz)-1)\,\mW_0$
and cannot leave $\operatorname{row}(\mW_0)$; a free latent or unconditioned control removes
$\vz$; a per-node hypernetwork frees both factors (\S\ref{app:design}); \method\ generates only
mixture coefficients over a shared learned low-rank basis. Tied-basis and learned-basis members
impose non-nested constraints (tied rows vs.\ free low-rank directions), so neither is expected
to dominate uniformly.

\subsection{\method: Structure-Conditioned Weight Generation}
\label{sec:swg}

We instantiate \eqref{eq:m3} by replacing the globally shared weight of \eqref{eq:m1} with a
node-specific effective weight generated from each node's precomputed structural fingerprint $\vz_v$
(e.g.\ degree, PageRank, random-walk statistics; see \S\ref{sec:fingerprint} for details).  Given
$\vz_v$, a lightweight generator predicts $r$ mixing coefficients that select from a shared set of
rank-one weight updates, and the resulting operator processes $\vx_v$ with no online message
passing.  Unlike \eqref{eq:m2}, $\vz_v$ here controls the student's linear operator directly;
unlike unrestricted hypernetworks, the generator produces only coefficients over a shared learned
basis, keeping per-node adaptation efficient while sharing adaptation directions across nodes.

An unrestricted realization of $\theta(\vz_v)$ could generate an entire predictor for every node, but doing so would be computationally expensive and would weaken sharing across related relational environments.  We instead assume that useful context-specific operators lie in a low-dimensional family around a globally shared base, mirroring the observation that the structural roles occupied by nodes in a large graph are few in number and recur across it~\cite{henderson2012rolx}.  Structural context therefore selects coefficients over a common set of adaptation directions rather than generating an independent weight matrix from scratch.
Concretely, we generate the node-specific effective weight as a shared base plus a low-rank, structure-conditioned update:
\begin{equation}
\label{eq:structlora}
\mW_{\mathrm{eff}}(v) \;=\; \mW_0 \;+\; \mU\,\mathrm{diag}\!\big(\vs(\vz_v)\big)\,\mV^\top,
\end{equation}
where $\mW_0\in\mathbb{R}^{o\times d}$ is a shared base weight, $\mU\in\mathbb{R}^{o\times r}$
and $\mV\in\mathbb{R}^{d\times r}$ form a shared learned low-rank basis of rank $r\ll\min(o,d)$,
and $\vs:\mathbb{R}^k\to\mathbb{R}^r$ is a small generator (a two-layer MLP) that maps the
structural fingerprint to a per-node coefficient vector.
Each paired column $(\mathbf u_i,\mathbf v_i)$ defines a rank-one adaptation direction $\mathbf u_i\mathbf v_i^\top$ shared across all instances, while $\vs(\vz_v)$ selects how strongly each direction is expressed in the relational environment of node $v$.  The basis therefore captures reusable patterns of predictor variation, and the fingerprint-dependent coefficients locate each node within this shared low-dimensional family.
The base $\mW_0$, basis $\mU,\mV$, and generator $\vs$ are trained jointly; \S\ref{app:design}
shows that this joint training carries real weight: freezing the shared components degrades accuracy sharply, and letting the generator emit both low-rank factors independently collapses to the structure-free baseline.

The layer output is computed \emph{without} materializing a per-node matrix:
\begin{equation}
\label{eq:structlora_fwd}
\mW_{\mathrm{eff}}(v)\,\vx_v \;=\; \mW_0\vx_v \;+\; \mU\big(\vs(\vz_v)\odot(\mV^\top\vx_v)\big).
\end{equation}
Since $\mW_0\vx_v$ and $\mV^\top\vx_v$ are both linear in $\vx_v$, they are realized jointly as
one stacked projection $[\mW_0;\mV^\top]\vx_v$.  Given the generator coefficients $\vs(\vz_v)$,
the adapted transformation therefore requires two batched linear projections and one element-wise product.
Excluding the one-time fingerprint preprocessing
and generator evaluation, the low-rank feature update adds $O(r(o+d))$ operations per node;
\S\ref{sec:exp} evaluates on graphs containing millions of nodes.
Each adapted student layer has its own base weight, low-rank basis, and coefficient generator.  \method\ adapts every student linear layer except the output head. We keep the output head globally shared so that all nodes use a common classifier, and for comparability across tables; \S\ref{app:design} shows that additionally adapting it gives a further gain. Biases in the adapted layers also remain shared.

\begin{figure}[t]
\centering
\includegraphics[width=\linewidth]{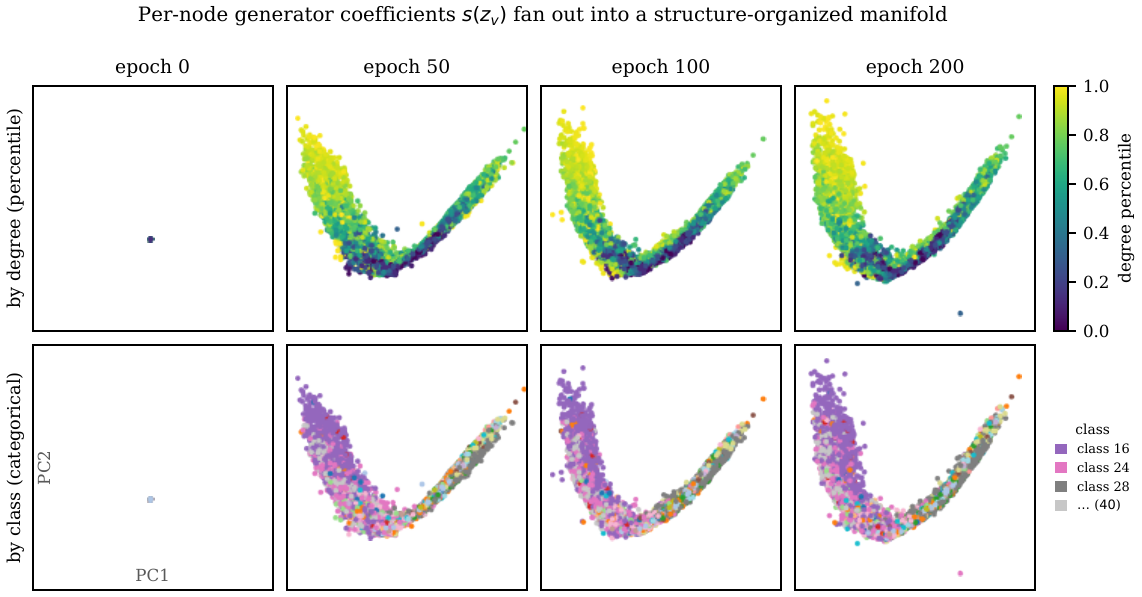}
\caption{2D PCA of the per-node generator output $\vs(\vz_v)$ on \textsc{ogbn-arxiv} at four
training epochs (shared axes), colored by degree percentile (top) and class label (bottom).
Organization by degree is consistent with sensitivity to structural context,
while class organization indicates task-aligned variation.}
\label{fig:manifold}
\vspace{-1em}
\end{figure}

\noindent\textbf{Baseline-preserving initialization.}
Zero-initializing the final layer of $\vs$ gives $\vs(\cdot)\equiv\mathbf{0}$, so the student is functionally equivalent to \eqref{eq:m1} at initialization and acquires structure-dependence only through learning (Fig.~\ref{fig:manifold}).

\noindent\textbf{Why generate weights instead of modulating activations.}
The tied-basis member \methoddiag\ applies $\vh_v=\operatorname{diag}(\gamma(\vz_v))\mW_0\allowbreak\vx_v+\beta(\vz_v)$, a row-wise rescaling of $\mW_0$; its linear action coincides with feature-wise modulation~\cite{perez2018film,dumoulin2018featurewise}. By contrast, $\mV$ in Eq.~\eqref{eq:structlora} is a free input projection, so $\mU\,\mathrm{diag}(\vs(\vz_v))\,\mV^\top$ need not equal $\mD\mW_0$ for any diagonal $\mD$; the effective weight can respond to input combinations no row-wise rescaling can produce (Prop.~\ref{prop:film}; empirical rank sweep in Fig.~\ref{fig:abrank}).

\subsection{Structural Fingerprints}
\label{sec:fingerprint}

For each node $v$, we precompute a low-dimensional structural fingerprint $\vz_v\in\mathbb{R}^k$.
Our default fingerprints are local permutation-equivariant topology statistics of the kind used to characterize a node's structural identity independently of its position in the graph~\cite{ribeiro2017struc2vec}: on smaller graphs these combine degree, clustering coefficient, and a random-walk structural encoding (RWSE)~\cite{dwivedi2022graph}; on large graphs, where dense random-walk powers are infeasible, they use only degree, neighborhood, and PageRank statistics computable by sparse matrix-vector products.  For transductive node classification, we additionally evaluate graph-specific positional signals such as LapPE, computable at this scale by a Lanczos solver.  Exact dimensions, variants, and precomputation
details are given in \S\ref{app:results}.  Fingerprints are computed once during graph
preprocessing and cached for repeated prediction: for transductive node classification this is
performed once for the full graph, while for graph classification it is performed independently for
each input graph before the message-passing-free forward pass.  For graph classification, where the model must transfer across separate graphs, we use local permutation-equivariant statistics rather than graph-specific positional coordinates, since eigenvector coordinates computed independently per graph are comparable across graphs only if the encoder is invariant to their sign and basis symmetries and stable under perturbations of the Laplacian~\cite{huang2024stable}.

\subsection{Training Objective}
\label{sec:objective}

The base weight $\mW_0$, basis $\mU,\mV$, and generator $\vs$ are trained jointly with the task
objective.  When a pretrained GNN teacher is available, we use
\begin{equation}
\label{eq:obj}
\mathcal{L} \;=\; (1-\lambda)\,\mathrm{CE}\big(\hat{\vy},\,y\big)\;+\;\lambda\,\mathrm{KL}\big(\mathbf{p}_T\,\|\,\hat{\mathbf{p}}\big),
\qquad \lambda\in[0,1].
\end{equation}
For node classification, where soft labels over unlabeled nodes are valuable, we use $\lambda{=}1$:
pure distillation, with hard labels supervising the student only indirectly through the teacher.
For graph classification, we use cross-entropy alone ($\lambda{=}0$) with no teacher or
distillation term (\S\ref{sec:exp_graph}).  In both settings, graph structure affects prediction
only through the cached fingerprint-conditioned weights; the student performs no message passing and
receives no propagated node features.  Although node-classification supervision comes from a
message-passing teacher, the teacher-free graph-classification setting confirms that
structure-conditioned weight generation is not intrinsically tied to distillation.

\subsection{Extending to Graph Classification}
\label{sec:graph_level}

\method\ is a general structure-conditioned MLP architecture; GNN-to-MLP distillation for node
classification is its primary evaluated application, and \S\ref{sec:objective} shows that the same
mechanism extends to the teacher-free setting.  For graph classification,
Eq.~\eqref{eq:structlora} generates each atom's weights from its own local fingerprint $\vz_v$,
unchanged.  An atom encoder and node-wise MLP produce per-atom outputs with \emph{no} message
passing, mean-pooled into a graph prediction and trained end-to-end with graph-level cross-entropy.

\noindent\textbf{A permutation-invariant alternative.}
When per-node fingerprints are unavailable or undesired, the same construction can instead condition
on a single permutation-invariant fingerprint $\vz_\gG$ for the whole graph, formed from the low
end of the normalized-Laplacian spectrum, degree-distribution statistics, and graph size.  For each
adapted layer $\ell$, the graph-level effective weight is
\begin{equation}
\label{eq:graphgen}
\mW^{(\ell)}_{\mathrm{eff}}(\gG) \;=\; \mW^{(\ell)}_0 \;+\;
\mU^{(\ell)}\,\mathrm{diag}\!\big(\vs^{(\ell)}(\vz_\gG)\big)\,\mV^{(\ell)\top},
\end{equation}
applying the same graph-conditioned weight to every atom at each adapted layer.  Both constructions
inherit symmetry from their conditioning signals: the per-node model is permutation-equivariant
whenever $\vz_v$ is equivariant, while the per-graph model is invariant when $\vz_\gG$ is
invariant (Prop.~\ref{prop:equiv}).
We use per-node generation as \method's primary graph-classification instantiation and
evaluate the per-graph variant as a matched comparison in \S\ref{sec:exp_graph}.

\subsection{Theoretical Analysis}
\label{sec:theory}

We characterize the model family induced by structure‑conditioned weight generation through three properties that directly support our design: (1) symmetry inheritance from the structural fingerprint, (2) operator variation beyond row‑wise activation modulation, and (3) approximability under a shared low‑rank basis. These are formal properties of the parameterisation rather than guarantees of optimisation, generalisation, or superiority over arbitrary activation-conditioning networks. They formalize the inductive bias of \method: structural context controls the feature-processing operator itself, and a shared, low-rank weight space enables efficient adaptation across nodes.

\begin{definition}[Permutation equivariance and invariance]
\label{def:perm}
Let $\pi$ be a node relabeling and $\pi\gG$ the relabeled graph with permuted features.
A node‑level predictor is \emph{permutation‑equivariant} if
$\hat{\vy}_{\pi(v)}(\pi\gG)\allowbreak=\hat{\vy}_v(\gG)$ for all $\pi,v$;
a graph‑level predictor is \emph{permutation‑invariant} if
$\hat{\vy}(\pi\gG)=\hat{\vy}(\gG)$ for all $\pi$.
\end{definition}

\begin{proposition}[Symmetry inheritance]
\label{prop:equiv}
If the node fingerprint $\vz_v$ is permutation-equivariant, then \method's predictions are permutation-equivariant. If the graph fingerprint $\vz_\gG$ is permutation-invariant, then every graph-conditioned layer satisfies $\mW_{\mathrm{eff}}^{(\ell)}(\pi\gG)=\mW_{\mathrm{eff}}^{(\ell)}(\gG)$, and the mean-pooled graph prediction is invariant. (Proof in App.~\ref{app:proof_equiv}.)
\end{proposition}

\noindent\textit{Remark.} Raw LapPE is not covered by this guarantee: eigenvectors carry sign and basis ambiguities~\cite{lim2023signnet}. We use it only for transductive node classification, where a single eigendecomposition of one fixed graph is reused; transferable settings use equivariant or invariant fingerprints, and covering raw LapPE would require a sign- and basis-invariant encoder~\cite{lim2023signnet,wang2022equivariant}.

\medskip
\noindent\textbf{Operator variation beyond row‑wise modulation.}
A central motivation for weight-space adaptation is that it can alter the linear operator beyond the specific row-wise activation-modulation family used in our experiments. Consider \methoddiag, which applies $\vh_v\leftarrow\gamma(\vz_v)\odot\vh_v+\beta(\vz_v)$ after a base linear map. Its context‑dependent linear perturbations lie in
\[
\mathcal{T}_{\mathrm{diag}}(\mW_0)=\{\mD\mW_0:\mD \text{ diagonal}\},
\]
i.e., only independent row scalings of the base weight matrix $\mW_0$.

In contrast, \method's perturbations belong to
\[
\mathcal{T}_{\method}(\mU,\mV)=\operatorname{span}\{\mathbf{u}_i\mathbf{v}_i^\top\}_{i=1}^{r},
\]
the span of the rank‑one basis matrices. This family can include perturbations that cannot be represented by independent row-wise rescaling of the base weight.

\begin{proposition}[Weight generation escapes the tied row-rescaling family]
\label{prop:film}
If there exists an index $i$ such that $\mathbf{u}_i\mathbf{v}_i^\top\notin\mathcal{T}_{\mathrm{diag}}(\mW_0)$, then $\mathcal{T}_{\method}(\mU,\mV)\allowbreak\not\subseteq\mathcal{T}_{\mathrm{diag}}(\mW_0)$. Consequently, some effective weights $\mW_{\mathrm{eff}}(v)=\mW_0+\mU\operatorname{diag}(\va)\mV^\top$ cannot be expressed as $\operatorname{diag}(\gamma)\mW_0$ for any $\gamma$. (Proof in App.~\ref{app:proof_film}.)
\end{proposition}

\noindent\textit{Why this matters.} The statement is about parameterisation, not universal expressive power (an MLP over $[\vx,\vz]$ can already approximate any function): it gives an architectural mechanism for per-node operator variation beyond row-wise rescaling, tested empirically in Figs.~\ref{fig:abrank} and~\ref{fig:weightspace_schematic}.

\medskip
\noindent\textbf{Shared‑basis approximability.}
The method's core assumption is that predictors optimal for different structural contexts vary along a small number of shared directions in weight space. To formalize this, let $\mW^*(\vz)$ be a \emph{fingerprint‑sufficient} target family: instances with the same fingerprint share the same optimal weight matrix. Assume that $\mathcal{Z}$ is compact and that there exist a base weight $\mW_0$, a shared basis $\mU,\mV$ of rank $r$, and a continuous coefficient map $\mathbf{a}:\mathcal{Z}\to\mathbb{R}^r$ such that
\begin{equation}
\sup_{\vz\in\mathcal{Z}}\big\|\mW^*(\vz)-\big(\mW_0+\mU\operatorname{diag}(\mathbf{a}(\vz))\mV^\top\big)\big\|_F \leq \delta_r .
\label{eq:shared_basis_assumption}
\end{equation}
Here $\delta_r$ measures how well the target weight family can be captured by a common rank‑$r$ affine subspace. This assumption directly reflects the \method\ architecture: a single shared basis is used for all nodes, and structural information only selects the coefficients.

\begin{corollary}[Approximation under a shared weight basis]
\label{cor:mapping}
Under Eq.~\eqref{eq:shared_basis_assumption}, with compact $\mathcal{Z}$ and per-instance losses uniformly $L$-Lipschitz in the layer weight on a compact set containing the target and generated weights, for any $\eta>0$ there exists a two-layer generator $\vs$ such that for every node $v$,
\[
\ell_v(\mW_0+\mU\operatorname{diag}(\vs(\vz_v))\mV^\top) \leq \ell_v(\mW^*(\vz_v)) + L(\delta_r+\eta).
\]
\end{corollary}

\noindent\textit{Interpretation.} The bound decomposes into shared-basis modelling error $\delta_r$ and generator error $\eta$; it does \emph{not} guarantee small $\delta_r$ at practical ranks or provide a generalization bound. Our rank sweeps and fingerprint-swap experiments supply indirect evidence rather than a direct estimate of $\delta_r$. Vectorising Eq.~\eqref{eq:structlora} gives $\operatorname{vec}(\mW_{\mathrm{eff}}(v))=\boldsymbol{\omega}_0+\mB\,\vs(\vz_v)$, matching a mapping network~\cite{sen2026mapping} but with $\mB$ restricted to vectorised rank-one factors and $\vs$ driven by structure rather than a free latent.

\section{Experiments}
\label{sec:exp}

Our experiments test four claims implied by viewing structural context as a source of model
adaptation. \textbf{RQ1} asks how the effectiveness of the same structural signal changes when it
is appended to the input, used for row-wise activation modulation, or used for weight adaptation. \textbf{RQ2} asks whether any gain
comes from meaningful conditioning rather than additional capacity, low-rank parameterization, or
weight generation alone. \textbf{RQ3} asks whether the learned weight space forms a shared but
context-specific family of predictors rather than applying a common shift to every instance.
\textbf{RQ4} asks when this principle improves graph prediction, whether it transfers beyond
node-level distillation, and what limits it. We emphasize at the outset that the answer to RQ1 is
\emph{signal-dependent}: no single conditioning route dominates every representation, and our
strongest positive results for weight adaptation appear for raw positional signals on the two OGB
graphs (Table~\ref{tab:signal}). Section~\ref{sec:exp_setup} first fixes the protocols and
matched controls. We report mean $\pm$ standard deviation over $10$ runs unless a table states otherwise.

\begin{table}[h]
\centering
\caption{Mechanism ladder on \textsc{ogbn-arxiv} under ho-full evaluation (\%). The axis swept is
\emph{how} structural information enters the model, \eqref{eq:m1}--\eqref{eq:m3}. M3 includes variants of \method.}
\label{tab:ladder}
\small
\begin{tabular}{lcc}
\toprule
Variant & Acc. & $\Delta$ vs. GLNN \\
\midrule
\multicolumn{3}{@{}l}{\itshape \eqref{eq:m1} structure-free} \\
GLNN (plain MLP $+$ KD) & 58.98$\pm$0.53 & 0.00 \\
\addlinespace[2pt]
\multicolumn{3}{@{}l}{\itshape \eqref{eq:m2} structure as input} \\
\;+ concatenate $\vz_v$ & 63.46$\pm$0.30 & +4.48 \\
\addlinespace[2pt]
\multicolumn{3}{@{}l}{\itshape \eqref{eq:m3} structure as adaptation context} \\
\multicolumn{3}{@{}l}{\quad\emph{without conditioning (controls)}} \\
\;free latent (Mapping-Network form) & 52.22$\pm$0.39 & $-6.76$ \\
\;unconditioned low-rank control ($UV^\top$, $r32$) & 58.05$\pm$0.48 & $-0.93$ \\
\multicolumn{3}{@{}l}{\quad\emph{with conditioning}} \\
\;\methoddiag\ (tied basis) & 63.84$\pm$0.22 & +4.86 \\
\;\method\ (learned basis, per-node) & \textbf{64.08$\pm$0.49} & \textbf{+5.10} \\
\bottomrule
\end{tabular}
\vspace{-1em}
\end{table}

\subsection{Experimental Setup}
\label{sec:exp_setup}

\noindent\textbf{Node classification.}
We use the five CPF benchmarks~\cite{yang2021extract} (\textsc{Cora}, \textsc{Citeseer},
\textsc{Pubmed}, \textsc{A-computer}, \textsc{A-photo}) and two OGB graphs~\cite{hu2020open}
(\textsc{ogbn-arxiv}, \textsc{ogbn-products}) under GLNN's~\cite{zhang2022graphless} splits,
released GraphSAGE teachers, and MLP backbones (App.~\ref{app:hyper}). We adopt GLNN's held-out
protocol: a random $20\%$ of test nodes is hidden from teacher and student training, and the teacher
sees only the observed subgraph, so held-out nodes and their edges never enter the soft labels.
Two settings differ in the graph used to compute fingerprints. \textbf{ho-full}
(held-out/full-structure) uses the full graph, so held-out-node edges enter the fingerprints, and a
training node's fingerprint may see held-out edges; this is \emph{structure-transductive}, and we
use it for the mechanism study, not as strict inductive generalisation. \textbf{ho-obs}
(held-out/observed-structure) uses the observed graph alone (fingerprints extended through observed
edges) and is reported in the system-level comparisons. Main runs use pure soft-label distillation
($\lambda{=}1$), the standard setting in this line, fixed across datasets to keep a shared
hyperparameter out of the matched comparisons.

\noindent\textbf{Matched mechanism controls.}
The central comparisons keep teacher, split, backbone, and structural signal fixed. GLNN omits the
fingerprint; \emph{Low-rank} replaces the shared linear layer by a rank-$r$ factorization
$\mU\mV^\top$ (no base weight, no conditioning) at \method's adaptation rank, so it tests whether
low-rank reparameterisation alone changes performance rather than ``unconditioned adaptation on top
of $\mW_0$''; \emph{\methoddiag} uses the fingerprint to modulate the first hidden activation; \method\
uses the same fingerprint to generate Eq.~\eqref{eq:structlora}. Input concatenation is evaluated
with the same signals in Table~\ref{tab:signal}. Ranks, widths, and tuning ranges are in
App.~\ref{app:hyper}; \method's ranks are fixed by benchmark family rather than tuned per dataset.
Because \methoddiag{} modulates one activation while \method\ adapts all hidden linear layers, we do not
attribute gains specifically to weight-space conditioning from the \methoddiag{} comparison alone.

\noindent\textbf{Graph classification.}
We use \textsc{ogbg-molhiv} with the scaffold split and five TU benchmarks with nested
cross-validation. All graph-classification variants train from task labels with cross-entropy only.
An edge-aware GIN is included as a same-protocol message-passing reference, not as a distillation
requirement. Complete protocols appear with the corresponding tables and in Appendix~\ref{app:graphcls}.

\begin{table}[h]
\centering
\footnotesize
\caption{Changing the structural signal and its role (\method rank $32$ on \textsc{arxiv}, $64$ on
\textsc{products}, \%). Parenthesized values are signal dimensions; $\Delta$ is the ho-full gain
over GLNN.}
\label{tab:signal}
\resizebox{\columnwidth}{!}{%
\begin{tabular}{@{}llcccccc@{}}
\toprule
&& \multicolumn{3}{c}{\textsc{ogbn-arxiv}} & \multicolumn{3}{c}{\textsc{ogbn-products}} \\
\cmidrule(lr){3-5}\cmidrule(l){6-8}
Signal (dim) & Used as & tran & ho-full & $\Delta$ & tran & ho-full & $\Delta$ \\
\midrule
\emph{Teacher} (SAGE) & message passing & 71.08$\pm$0.42 & 70.52$\pm$0.36 & & 78.15$\pm$0.39 & 78.18$\pm$0.26 & \\
none (GLNN) & -- & 64.81$\pm$0.28 & 58.98$\pm$0.53 & 0.0 & 64.60$\pm$0.34 & 64.16$\pm$0.19 & 0.0 \\
\midrule
\multicolumn{8}{@{}l}{\itshape node-wise spectral structural summaries} \\
HKS (8) & generation & 66.11$\pm$0.36 & 59.51$\pm$0.63 & +0.5 & 63.70$\pm$0.30 & 63.15$\pm$0.26 & $-1.0$ \\
RWSE (8) & generation & 66.04$\pm$0.25 & 59.75$\pm$0.44 & +0.8 & 63.68$\pm$0.46 & 63.15$\pm$0.29 & $-1.0$ \\
\addlinespace[2.5pt]
\multicolumn{8}{@{}l}{\itshape positional and local signals, used to generate weights} \\
degree/PageRank (7) & generation & 68.87$\pm$0.40 & 64.08$\pm$0.49 & +5.1 & 70.52$\pm$0.30 & 69.98$\pm$0.30 & +5.8 \\
LapPE (8) & generation & 70.69$\pm$0.36 & 67.96$\pm$0.56 & +9.0 & 65.05$\pm$0.32 & 64.51$\pm$0.24 & +0.4 \\
LapPE (32) & generation & 71.47$\pm$0.31 & 69.83$\pm$0.33 & +10.9 & 73.55$\pm$0.46 & 73.07$\pm$0.40 & +8.9 \\
DeepWalk (128) & generation & 71.56$\pm$0.30 & 69.79$\pm$0.38 & +10.8 & 78.93$\pm$0.44 & 78.63$\pm$0.35 & +14.5 \\
\addlinespace[2.5pt]
\multicolumn{8}{@{}l}{\itshape the same signals in other roles} \\
LapPE (32) & modulation & 71.54$\pm$0.33 & 70.05$\pm$0.23 & +11.1 & 72.71$\pm$0.40 & 72.66$\pm$0.41 & +8.5 \\
LapPE (32) & input concat & 68.94$\pm$0.23 & 59.27$\pm$0.86 & +0.3 & 65.00$\pm$0.52 & 65.69$\pm$0.54 & +1.5 \\
DeepWalk (128) & input concat & 71.72$\pm$0.30 & 70.48$\pm$0.42 & +11.5 & 79.01$\pm$0.43 & 78.83$\pm$0.27 & +14.7 \\
\addlinespace[2.5pt]
LapPE weights + DW inputs & composition & \textbf{71.77$\pm$0.31} & \textbf{70.80$\pm$0.19} & \textbf{+11.8} & \textbf{79.44$\pm$0.36} & \textbf{79.23$\pm$0.30} & \textbf{+15.1} \\
\bottomrule
\end{tabular}%
}
\vspace{-1em}
\end{table}

\subsection{Structural Context as Input, Modulation, or Weight Adaptation}
\label{sec:exp_signal}

RQ1 compares different roles for the same structural context. Table~\ref{tab:ladder} first fixes the
default fingerprint on \textsc{ogbn-arxiv}. Appending the fingerprint already recovers much of the
benefit, confirming that the signal itself is useful. Modulation and weight generation improve
further, although their differences under this inexpensive fingerprint are comparable to seed
variation. The stronger separation appears when the signal is informative enough to expose the
choice of route.

Table~\ref{tab:signal} then varies both signal and route on the two OGB graphs. Invariant summaries such as the heat-kernel signature~\cite{sun2009hks} and RWSE~\cite{dwivedi2022graph} carry limited node-specific information and barely improve the student. Raw
positional signals are substantially stronger. With LapPE, input concatenation remains close to the
structure-free student under held-out-node evaluation, while modulation and generation retain most
of the available gain. On \textsc{ogbn-products}, weight generation is numerically higher than
activation modulation, though within a few tenths of a percentage point. This contrast is not universal to every representation: DeepWalk has already learned a
smooth graph-specific embedding and works similarly as an input or a conditioning signal. The
composition row shows that the two roles are complementary rather than mutually exclusive.

\begin{tcolorbox}[colback=gray!5, colframe=gray!50, boxrule=0.5pt, left=2pt, right=2pt, top=2pt, bottom=2pt]
\textbf{RQ1 answer, and a route-selection heuristic.} No route dominates every signal: generated
weights are most clearly advantageous for raw LapPE on the two OGB graphs (where input concat stays
near the structure-free student), row-wise modulation is competitive for LapPE, and DeepWalk is
numerically indistinguishable across routes within seed variation (arxiv ho-full $70.48\pm0.42$
input vs.\ $69.79\pm0.38$ generation; products $78.83$ vs.\ $78.63$). The pattern suggests a simple
heuristic: fingerprints with symmetry or dynamic-range ambiguities the input pathway cannot resolve
(raw LapPE, degree/PageRank) benefit from weight adaptation; already-trained smooth embeddings
(DeepWalk) do not; permutation-invariant summaries (HKS, RWSE) at their tested dimensions carry
little node-specific information, and neither route helps. These matched results use
\emph{ho-full}, so they characterise how full-graph cached structure is used, not strict unseen-node
generalisation (see \emph{ho-obs} in Table~\ref{tab:ogb_systems}).
\end{tcolorbox}

\begin{figure}[h]
\centering
\begin{subfigure}[b]{0.47\linewidth}
  \centering
  \includegraphics[width=\linewidth]{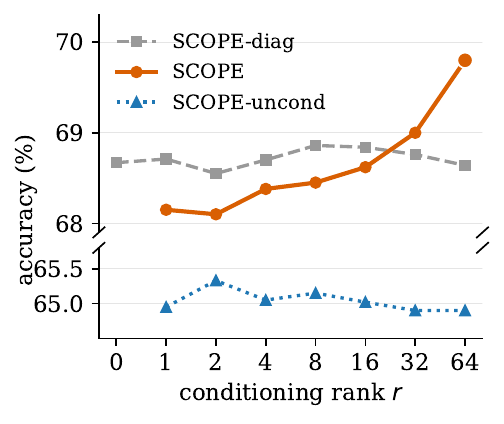}
  \caption{Adaptation rank.}
  \label{fig:abrank}
\end{subfigure}
\hfill
\begin{subfigure}[b]{0.47\linewidth}
  \centering
  \raisebox{2mm}{\includegraphics[width=\linewidth]{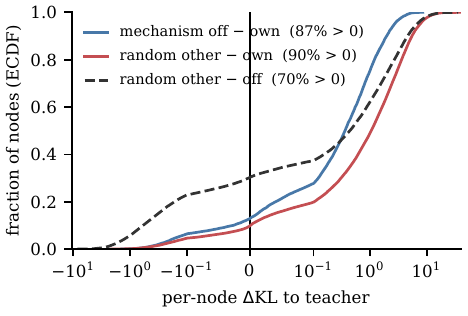}}
  \caption{Per-node functional test.}
  \label{fig:func_ecdf}
\end{subfigure}
\caption{Rank behaviour and node-specificity on \textsc{ogbn-arxiv}.
(a) Transductive accuracy (\%) versus adaptation rank.
(b) ECDF of per-node $\Delta$KL to the teacher (\method\ $r32$); positive favours a node's own
fingerprint.}
\label{fig:rank_func}
\Description{Left: accuracy versus conditioning rank for Diag and \method. Right: empirical CDF of per-node KL differences under own, swapped, and disabled fingerprints.}
\vspace{-1em}
\end{figure}

\subsection{Unconditioned Low-Rank Alternatives Do Not Reproduce the Gain}
\label{sec:exp_ablation}

RQ2 tests whether the gain can be reproduced by the unconditioned low-rank alternatives
considered in this work. The first three rows of Table~\ref{tab:ladder} show that neither a free
latent nor the tested unconditioned low-rank parameterisation improves the shared student. The rank-matched Low-rank control in
Table~\ref{tab:node_main} gives the same result across datasets.
A factor-emitting conditional hypernetwork, which generates both low-rank factors per node, also fails to improve over the structure-free baseline (Appendix~\ref{app:design}).
Together, these results are consistent with the
benefit of generating coefficients over a shared learned basis, rather than proving that this
particular parameterisation is necessary.

\noindent\textbf{Rank-dependent behaviour of the tested conditioning architectures.}
Figure~\ref{fig:abrank} varies the adaptation rank under \methoddiag{} and generated weights. Increasing
rank has little effect on the tested first-layer \methoddiag{} architecture, whose linear action remains a
rescaling of the base rows. Under \method, rank opens additional adaptation directions and improves
performance until saturation. A deeper coefficient generator does not provide a comparable
improvement, so within the tested architectures the useful capacity appears to lie in the weight
subspace rather than in a larger conditioning network. The two arms differ in mounting depth, so this characterises their rank behaviour rather than isolating weight space (Section~\ref{sec:exp_setup}).

\begin{tcolorbox}[colback=gray!5, colframe=gray!50, boxrule=0.5pt, left=2pt, right=2pt, top=2pt, bottom=2pt]
Within the tested controls, unconditioned weight generation is not beneficial: these results support the view that a structural signal selecting how the shared weight space is used is what recovers the improvement.
\end{tcolorbox}

\subsection{A Structure-Organized Family of Predictors}
\label{sec:exp_geometry}

RQ3 tests whether the generator learns instance-specific adaptation rather than a common correction.
The strongest evidence is functional. For each test node, we compare its prediction using its own
fingerprint, a randomly selected other node's fingerprint, and the base model with adaptation
disabled. Across ranks, the correct fingerprint helps most nodes and outperforms the swapped one for
roughly nine out of ten nodes (Table~\ref{tab:func_main}). The wrong fingerprint is often worse than
turning the mechanism off, so the generated update is addressed to the structural context of the
node rather than acting as a shared bias.

\begin{table}[h]
\centering
\footnotesize
\caption{Functional node-specificity on \textsc{ogbn-arxiv}. Struct-gain is
$\mathrm{KL}_{\mathrm{off}}-\mathrm{KL}_{\mathrm{own}}$ and node-spec. is
$\mathrm{KL}_{\mathrm{swap}}-\mathrm{KL}_{\mathrm{own}}$.}
\label{tab:func_main}
\setlength{\tabcolsep}{5pt}
\begin{tabular}{lccccc}
\toprule
Config & $\mathrm{KL}_{\mathrm{own}}$ & struct-gain & frac. helped & node-spec. & frac. own better \\
\midrule
$r16$ & 0.388 & 0.647 & 0.854 & 1.896 & 0.891 \\
$r32$ & 0.358 & 0.691 & 0.870 & 2.041 & 0.900 \\
$r64$ & 0.315 & 0.726 & 0.888 & 2.074 & 0.915 \\
$r32$ thick & 0.329 & 0.764 & 0.893 & 2.366 & 0.922 \\
\bottomrule
\end{tabular}
\end{table}

\begin{figure}[h]
\centering
\includegraphics[width=\linewidth]{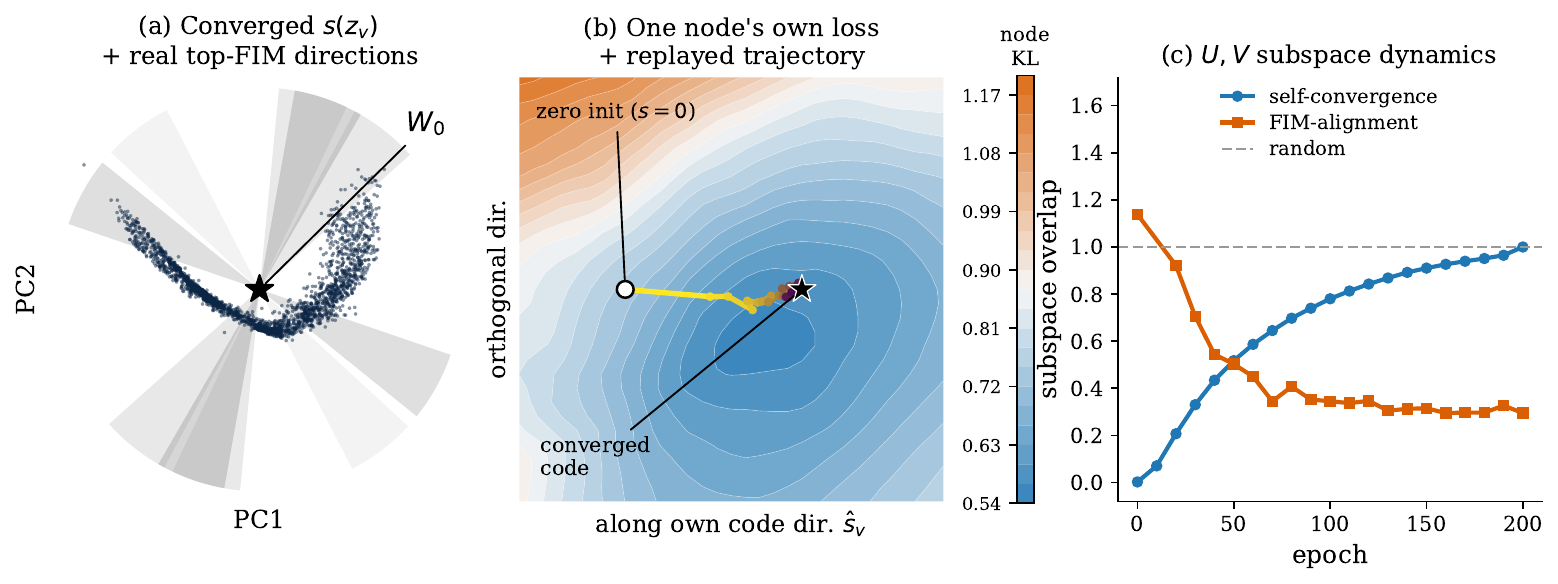}
\caption{Measured weight-space diagnostics on \textsc{ogbn-arxiv}. (a) Per-node generator outputs
with chart-restricted Fisher directions. (b) A node's replayed optimization path. (c) Convergence
of the structure-reachable subspace.}
\label{fig:weightspace_schematic}
\vspace{-1em}
\end{figure}

The learned updates are geometrically distinct from simple rescaling: nontrivial in magnitude, nearly Frobenius-orthogonal to the base weight, and mostly outside its dominant rank-matched singular subspace. During training, generator outputs spread from the origin into a stable \emph{chart} organized by structural role (Fig.~\ref{fig:manifold}). Empirical-Fisher diagnostics (Fig.~\ref{fig:weightspace_schematic}; App.~\ref{app:fim}) further show this chart develops below-random overlap with the leading Fisher eigenspace.

\begin{tcolorbox}[colback=gray!5, colframe=gray!50, boxrule=0.5pt, left=2pt, right=2pt, top=2pt, bottom=2pt]
One shared generator produces a related family of predictors whose differences are associated with
structural context. The functional swap test provides the primary evidence for RQ3; the parameter-space
visualisations are descriptive and do not by themselves establish a causal explanation.
\end{tcolorbox}

\subsection{Utility, Transfer, and Limitations}
\label{sec:exp_scope}
\label{sec:exp_node}
\label{sec:exp_graph}

\begin{table*}[t]
\centering
\caption{Node classification accuracy. Graph-based and positional rows are task-level references
rather than matched mechanism controls.}
\label{tab:node_main}
\scriptsize
\setlength{\tabcolsep}{1.5pt}
\begin{tabular}{lccccccc|ccccccc}
\toprule
& \multicolumn{7}{c|}{Transductive} & \multicolumn{7}{c}{Held-out nodes (ho-full)} \\
Method & Cora & Citeseer & Pubmed & A-computer & A-photo & ogbn-arxiv & ogbn-products & Cora & Citeseer & Pubmed & A-computer & A-photo & ogbn-arxiv & ogbn-products \\
\midrule
\multicolumn{15}{l}{\emph{Graph-based task references}} \\
Teacher (SAGE) & 80.49$\pm$1.90 & 70.23$\pm$2.15 & 75.42$\pm$1.85 & 82.11$\pm$1.88 & 90.71$\pm$1.26 & 71.08$\pm$0.42 & 78.15$\pm$0.39 & 81.24$\pm$1.91 & 70.00$\pm$2.91 & 75.37$\pm$2.74 & 82.91$\pm$1.56 & 91.04$\pm$1.21 & 70.52$\pm$0.36 & 78.18$\pm$0.26 \\
SGC & 80.81$\pm$1.37 & 70.91$\pm$2.36 & 74.53$\pm$1.91 & 83.74$\pm$1.58 & 91.13$\pm$1.12 & 69.31$\pm$0.31 & 68.26$\pm$0.15 & 80.33$\pm$2.13 & 70.83$\pm$3.02 & 72.94$\pm$2.63 & 83.58$\pm$1.84 & 91.31$\pm$0.81 & 69.07$\pm$0.69 & 68.45$\pm$0.16 \\
\midrule
\multicolumn{15}{l}{\emph{Learned positional-input reference (CE, no teacher)}} \\
DW+MLP & 75.50$\pm$1.25 & 62.35$\pm$2.80 & 67.98$\pm$2.75 & 76.08$\pm$1.86 & 79.87$\pm$1.51 & 69.45$\pm$0.25 & 78.06$\pm$0.23 & 71.38$\pm$2.78 & 63.92$\pm$3.62 & 65.34$\pm$2.96 & 80.30$\pm$1.43 & 81.13$\pm$1.72 & 69.81$\pm$0.35 & 78.88$\pm$0.43 \\
\midrule
\multicolumn{15}{l}{\emph{Matched shared-backbone controls (KD)}} \\
GLNN & 79.98$\pm$1.92 & 71.03$\pm$2.22 & 75.70$\pm$2.04 & 82.61$\pm$1.91 & 91.69$\pm$1.28 & 64.81$\pm$0.28 & 64.60$\pm$0.34 & \textbf{74.15$\pm$1.81} & 68.78$\pm$3.13 & 75.11$\pm$2.59 & 81.06$\pm$1.45 & 89.22$\pm$1.03 & 58.98$\pm$0.53 & 64.16$\pm$0.19 \\
Low-rank & \textbf{80.54$\pm$2.15} & 70.28$\pm$2.24 & 75.73$\pm$1.98 & 81.88$\pm$1.99 & 91.91$\pm$1.09 & 60.08$\pm$0.30 & 63.90$\pm$0.28 & 72.15$\pm$2.29 & 67.35$\pm$2.91 & 75.02$\pm$2.69 & 80.34$\pm$1.41 & 87.48$\pm$0.73 & 58.05$\pm$0.48 & 63.41$\pm$0.28 \\
\methoddiag & 80.43$\pm$2.53 & 70.76$\pm$1.60 & 76.05$\pm$2.20 & 82.28$\pm$2.01 & 91.45$\pm$0.98 & 68.51$\pm$0.44 & 69.75$\pm$0.34 & 73.00$\pm$2.54 & \textbf{69.97$\pm$2.12} & 75.48$\pm$2.43 & 80.88$\pm$1.48 & 89.39$\pm$1.42 & 63.84$\pm$0.22 & 69.24$\pm$0.30 \\
\method & 80.44$\pm$1.94 & \textbf{71.36$\pm$2.12} & 75.99$\pm$2.14 & 82.04$\pm$1.69 & \textbf{92.06$\pm$0.93} & 68.87$\pm$0.40 & 70.52$\pm$0.30 & 73.75$\pm$1.58 & 68.92$\pm$3.95 & 74.83$\pm$2.35 & 81.18$\pm$1.38 & 89.66$\pm$1.17 & 64.08$\pm$0.49 & 69.98$\pm$0.30 \\
+LapPE & 80.10$\pm$2.01 & 71.15$\pm$2.07 & \textbf{76.21$\pm$2.21} & \textbf{82.75$\pm$1.78} & 91.56$\pm$0.99 & \textbf{71.47$\pm$0.31} & \textbf{73.55$\pm$0.46} & 73.89$\pm$2.43 & 69.14$\pm$3.11 & \textbf{75.64$\pm$2.61} & \textbf{83.24$\pm$1.58} & \textbf{91.52$\pm$1.09} & \textbf{69.83$\pm$0.33} & \textbf{73.07$\pm$0.40} \\
\midrule
$\Delta$(+LapPE$-$GLNN) & +0.1 & +0.1 & +0.5 & +0.1 & $-0.1$ & \textbf{+6.7} & \textbf{+8.9} & $-0.3$ & +0.4 & +0.5 & \textbf{+2.2} & \textbf{+2.3} & \textbf{+10.9} & \textbf{+8.9} \\
\bottomrule
\end{tabular}
\end{table*}

RQ4 tests where the mechanism is useful across graph benchmarks. Tables~\ref{tab:node_main} and~\ref{tab:ogb_systems} give two views under a shared backbone and protocol: the first is a matched-control comparison; the second pairs each published system with the \method\ variant using a comparable structural signal (TINED transfers teacher weights; NOSMOG adds a DeepWalk positional input; TINED+ combines both). Small CPF datasets show structure-conditioned weights close to the shared student (details in App.~\ref{app:cpf_systems}); OGB shows large separations, with both the default fingerprint and LapPE improving substantially. This dataset-family pattern is consistent with the motivating hypothesis that a fixed predictor becomes more restrictive as structural-context diversity grows, but graph size is confounded with label rate, domain, fingerprint, architecture, and rank and is not isolated as the causal factor. The ho-obs columns of Table~\ref{tab:ogb_systems} compute \method's fingerprints from the observed graph only, so no full-graph structure enters held-out predictions. Table~\ref{tab:ogb_systems} is meant to show consistency with, not universal superiority over, DeepWalk-input and teacher-transfer systems; the matched-signal isolation in Tables~\ref{tab:ladder}--\ref{tab:signal} remains our strongest evidence for the mechanism itself.

\begin{table}[t]
\centering
\footnotesize
\caption{System-level context on the two OGB graphs. External numbers are as published; \method\
uses our protocol under ho-obs (\S\ref{sec:exp_setup}). Bold marks the higher entry of a pair.}
\label{tab:ogb_systems}
\setlength{\tabcolsep}{3pt}
\resizebox{\columnwidth}{!}{%
\begin{tabular}{@{}llcccc@{}}
\toprule
& & \multicolumn{2}{c}{\textsc{ogbn-arxiv}} & \multicolumn{2}{c@{}}{\textsc{ogbn-products}} \\
\cmidrule(lr){3-4}\cmidrule(l){5-6}
Method & Structural route & tran & ho-obs & tran & ho-obs \\
\midrule
TINED~\cite{zhou2025tined} & teacher-weight transfer & 64.44$\pm$0.72 & 59.79$\pm$0.46 & 69.48$\pm$0.25 & 68.68$\pm$0.27 \\
\method & conditioned weights & \textbf{68.87$\pm$0.40} & \textbf{63.73$\pm$0.45} & \textbf{70.52$\pm$0.30} & \textbf{69.44$\pm$0.26} \\
\midrule
NOSMOG~\cite{tian2023nosmog} & DeepWalk input & \textbf{71.65$\pm$0.29} & \textbf{70.09$\pm$0.55} & 78.45$\pm$0.38 & 77.02$\pm$0.19 \\
\method + DeepWalk & DeepWalk-conditioned weights & 71.56$\pm$0.30 & 69.98$\pm$0.47 & \textbf{78.93$\pm$0.44} & \textbf{78.63$\pm$0.34} \\
\midrule
TINED+~\cite{zhou2025tined} & teacher transfer + DeepWalk input & 71.52$\pm$0.34 & 70.42$\pm$0.35 & 78.59$\pm$0.28 & \textbf{79.31$\pm$0.29} \\
\method + DW input & conditioned weights + DeepWalk input & \textbf{71.67$\pm$0.31} & \textbf{70.48$\pm$0.48} & \textbf{79.23$\pm$0.42} & 78.53$\pm$0.44 \\
\bottomrule
\end{tabular}%
}
\vspace{-0.8em}
\end{table}

\begin{figure}[h]
\centering
\begin{subfigure}[b]{0.47\linewidth}
  \centering
  \includegraphics[width=\linewidth]{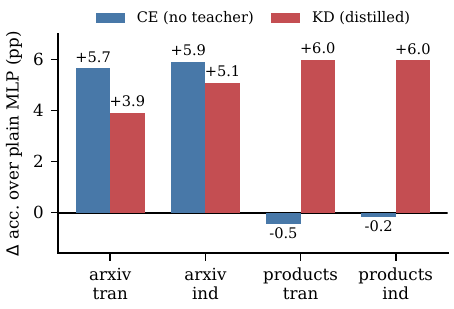}
  \caption{Teacher-free increment.}
  \label{fig:noteacher}
\end{subfigure}
\hfill
\begin{subfigure}[b]{0.47\linewidth}
  \centering
  \includegraphics[width=\linewidth]{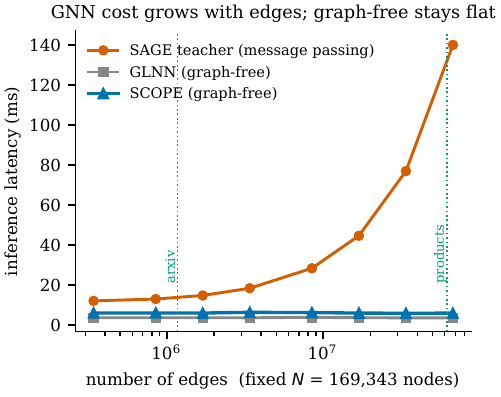}
  \caption{Prediction latency.}
  \label{fig:efficiency}
\end{subfigure}
\caption{Utility and cost on the two OGB graphs. (a) Increment from \method\ over the matching
plain MLP under CE and KD. (b) Prediction latency versus graph density; dotted lines mark the two
datasets.}
\label{fig:utility_cost}
\Description{Left: accuracy increment from \method with and without distillation on two OGB graphs. Right: inference latency versus graph density for message passing, a shared MLP, and \method.}
\end{figure}

\noindent\textbf{Transfer beyond node distillation.}
Table~\ref{tab:graph} applies the same mechanism to graph classification, trained from scratch with cross-entropy (no teacher, unlike graph-level GNN-to-MLP distillation~\cite{yao2024mugsi}). This is a different instantiation than the per-node distillation above: per-atom (Eq.~\ref{eq:structlora}, primary) and per-graph (Eq.~\ref{eq:graphgen}) variants share the low-rank generator but differ in conditioning granularity, so results should not be transported across the two settings.
On
\textsc{ogbg-molhiv}, local per-node conditioning is strongest, consistent with atoms in the same
molecule requiring different computations. On several TU datasets, a graph-level fingerprint or a
single graph-conditioned weight set already captures much of the target, while per-node adaptation
helps where local context matters. The result is not a universal replacement for message passing:
on \textsc{NCI1}, none of the fingerprints contains the fine chemical substructure needed by the
task, and GIN remains substantially stronger.

\begin{table}[t]
\centering

\caption{Graph classification accuracy. \textsc{ogbg-molhiv} reports ROC-AUC over seeds; the five TU columns report nested-cross-validation accuracy over folds. Generation and broadcast variants use rank~$64$.}
\label{tab:graph}
\begingroup
\scriptsize
\setlength{\tabcolsep}{2pt}
\renewcommand{\arraystretch}{1.05}
\begin{tabular}{@{}lr@{$\pm$}l@{\hspace{6pt}}r@{$\pm$}lr@{$\pm$}lr@{$\pm$}lr@{$\pm$}lr@{$\pm$}l@{}}
\toprule
 & \multicolumn{2}{c}{ROC-AUC} & \multicolumn{10}{c@{}}{10-fold accuracy} \\
\cmidrule(lr){2-3}\cmidrule(l){4-13}
Arm & \multicolumn{2}{c}{molhiv} & \multicolumn{2}{c}{PROTEINS} & \multicolumn{2}{c}{DD}
& \multicolumn{2}{c}{IMDB-B} & \multicolumn{2}{c}{REDDIT-B} & \multicolumn{2}{c@{}}{NCI1} \\
\midrule
\multicolumn{13}{@{}l}{\itshape no structural signal} \\
plain MLP & 68.3 & 0.7 & 60.6 & 2.2 & 70.3 & 3.8 & 70.2 & 3.6 & 62.8 & 1.6 & 57.1 & 1.4 \\
\addlinespace[2pt]
\multicolumn{13}{@{}l}{\itshape structure as input} \\
fingerprint-only\,$\vz_\gG$ & 70.4 & 0.6 & 74.2 & 3.8 & 76.2 & 2.9 & 70.1 & 1.6 & \textbf{83.3} & 2.2 & 70.0 & 2.4 \\
input-concat\,$\vz_v$ & 76.7 & 0.9 & 68.4 & 4.0 & 72.6 & 3.3 & 71.8 & 4.0 & 78.8 & 2.0 & 68.7 & 2.1 \\
\addlinespace[2pt]
\multicolumn{13}{@{}l}{\itshape structure as weights (ours)} \\
\method\ per-graph & 71.6 & 1.0 & \textbf{75.0} & 3.9 & \textbf{77.2} & 3.2 & 71.9 & 2.3 & 80.3 & 4.4 & 69.6 & 2.9 \\
\method\ per-node & \textbf{77.3} & 1.4 & 68.5 & 4.3 & 69.6 & 6.2 & \textbf{73.1} & 2.9 & 80.1 & 2.9 & 68.4 & 4.7 \\
\addlinespace[2pt]
\multicolumn{13}{@{}l}{\itshape message-passing reference} \\
GIN (edge-aware) & 74.6 & 1.8 & 74.4 & 3.4 & 73.2 & 5.2 & 68.7 & 3.5 & 81.2 & 2.2 & \textbf{81.4} & 1.6 \\
\bottomrule
\end{tabular}
\endgroup

\end{table}

All TU results use nested selection: the stopping epoch is chosen on a held-out validation split, a
fresh model is retrained for that duration, and the test fold is evaluated once per retrain. This
avoids the test-fold selection bias discussed by Errica et al.~\cite{errica2020fair}.

\phantomsection
\label{sec:exp_teacher}
\noindent\textbf{The mechanism does not require a teacher, but supervision still matters.}
Figure~\ref{fig:noteacher} compares the increment from weight adaptation with and without
distillation on the OGB graphs. On \textsc{ogbn-arxiv}, labels alone are sufficient to learn a useful
conditioning rule. On \textsc{ogbn-products}, the same mechanism benefits strongly from teacher outputs, consistent with the much lower labeled fraction. That is the regime motivating boosted multi-student GNN-to-MLP distillation~\cite{lu2024adagmlp}, although graph scale and split construction are confounded with label availability. The teacher-free graph-classification results provide the cleaner conclusion: distillation is one source of supervision for node classification, not a component of the weight generation.

\begin{tcolorbox}[colback=gray!5, colframe=gray!50, boxrule=0.5pt, left=2pt, right=2pt, top=2pt, bottom=2pt]
Under RQ4, structure-conditioned weights help most when the fingerprint exposes task-relevant
variation and hurt none, but cannot recover absent structural information and require enough
observed structure to build the fingerprint (strict cold start is out of scope).
Post-caching inference on \textsc{ogbn-products} runs in 101\,ms, compared with 412\,ms for the teacher;
latency is independent of graph density, and the learned parameter count is independent of graph size (Fig.~\ref{fig:efficiency}; App.~\ref{app:latency}).
\end{tcolorbox}

\section{Conclusion}
\label{sec:conclusion}
Our study reframes what ``using graph structure'' can mean for a predictor that performs no online
message passing: the same cached fingerprint can enter as an input feature, an activation
modulator, or a coordinate that selects a member of a shared family of effective predictors.
That the preferred route depends on the representation is not a caveat but the substantive
finding. It says the conditioning route is a modelling decision coupled to how much structural
work the fingerprint has already done: analytic positional coordinates reward being pushed deeper
into the operator, whereas already-trained embeddings do not. This offers a working hypothesis for
why prior graph-to-MLP work has converged on a mix of inputs, distillation targets, and
initialisations without a single winner---each route implicitly commits to how much structural
computation the predictor still has to do downstream.
Three gaps in the controlled study are natural next steps: whether a depth-matched \methoddiag\ arm would close the OGB separation; whether the
shared-basis approximation error remains small for non-topological structured domains such as
sequences, meshes, or molecular fragments; and how the coefficient generator behaves under
fingerprint distribution shift across graph families. Strict cold-start deployment~\cite{zheng2022cold}
and heterogeneous-relation transfer~\cite{liu2025hg2m} lie beyond the fingerprint-driven scope of
this work.

\bibliographystyle{ACM-Reference-Format}
\bibliography{references,weightspace_refs}

\appendix

\section{Supplementary Material}
\label{app:results}

\begin{figure}[h]
\centering
\includegraphics[width=\linewidth]{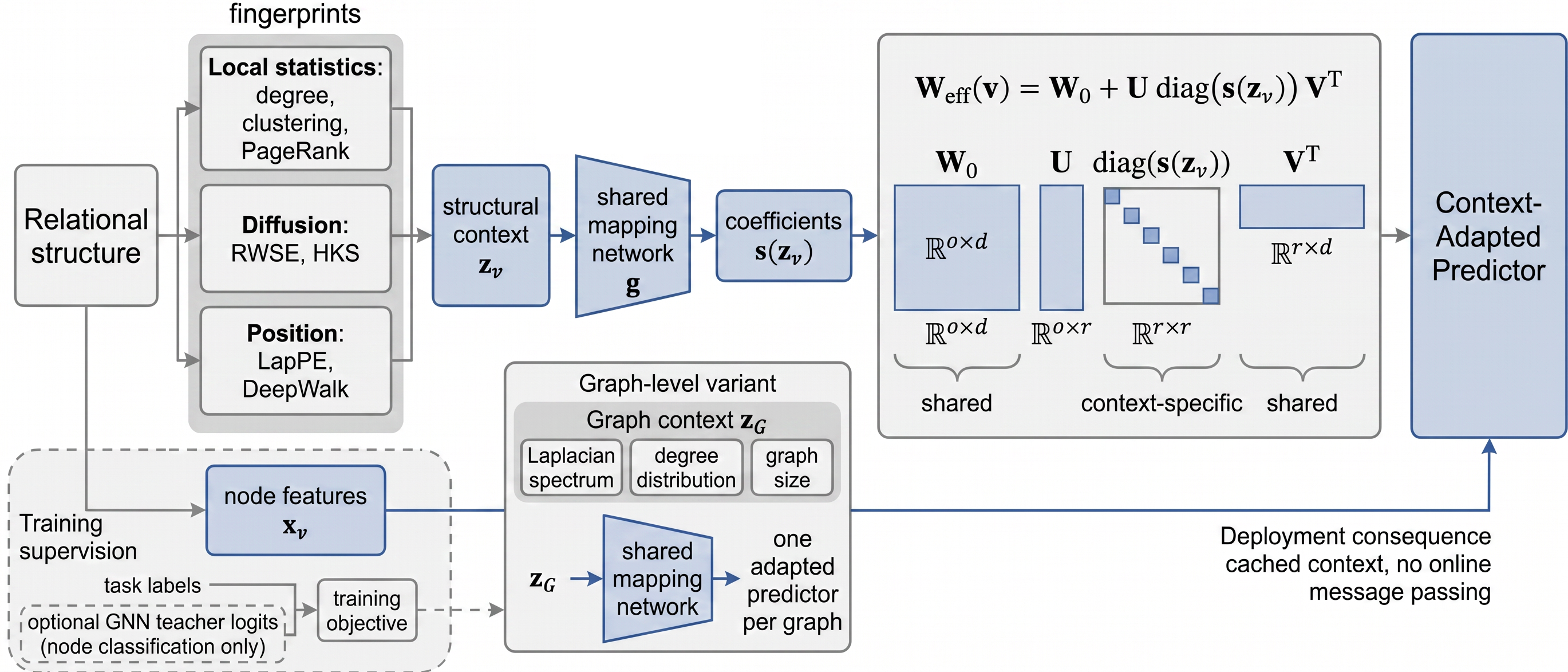}

\caption{\method{} overview: a cached structural fingerprint $\mathbf{z}_v$ is mapped to coefficients $\mathbf{s}(\mathbf{z}_v)$ that select a context-specific update to the shared base weight. The graph-level variant broadcasts one $\mathbf{z}_\mathcal{G}$ per graph.}

\label{fig:overview}
\vspace{-1em}
\end{figure}

All accuracies and AUCs are reported as percentages, matching the main text;
Node-classification results are reported under the
transductive (tran), held-out/full-structure (ho-full), and held-out/observed-structure (ho-obs;
see \S\ref{sec:exp_setup}) evaluations. Graph classification reports ROC-AUC on
\textsc{ogbg-molhiv} and accuracy on the TU datasets. Unless noted,
the GNN-to-MLP setup follows \cite{zhang2022graphless,tian2023nosmog}: the
student is a message-passing-free MLP, and the teacher is a GNN. The per-node structural fingerprint $z_v$ (degree, log-degree, neighbor-degree statistics, PageRank, and random-walk return probabilities) is cached once per dataset. The ``thick''
variant enriches $z_v$ with eigenvector centrality, neighbor PageRank,
neighbor-degree dispersion, and additional hops.
Figure~\ref{fig:overview} summarises the architecture.

\begin{table*}[!t]
\centering
\footnotesize
\caption{Node-classification datasets, architectures, and training hyperparameters.
lr/wd/dropout are the student's; $r$ is \method's rank, $k$ the fingerprint dimension.}
\label{tab:hyper}
\setlength{\tabcolsep}{4pt}
\begin{tabular}{lrrrrcccccc}
\toprule
Dataset & nodes & edges & feat.\ dim & classes & teacher & student & lr & wd & drop.\ & $r$ / $k$ \\
\midrule
Cora          & 2,485     & 5,069      & 1,433 & 7  & 2$\times$128 & 2$\times$128 & 0.010 & 5e-3 & 0.6 & 8 / 18 \\
Citeseer      & 2,110     & 3,668      & 3,703 & 6  & 2$\times$128 & 2$\times$128 & 0.010 & 1e-3 & 0.1 & 8 / 18 \\
Pubmed        & 19,717    & 44,324     & 500   & 3  & 2$\times$128 & 2$\times$128 & 0.005 & 1e-3 & 0.4 & 8 / 18 \\
A-computer    & 13,381    & 245,778    & 767   & 10 & 2$\times$128 & 2$\times$128 & 0.003 & 5e-3 & 0.1 & 8 / 18 \\
A-photo       & 7,487     & 119,043    & 745   & 8  & 2$\times$128 & 2$\times$128 & 0.002 & 1e-3 & 0.3 & 8 / 18 \\
ogbn-arxiv    & 169,343   & 1,166,243  & 128   & 40 & 3$\times$256 BN & 3$\times$256 BN & 0.010 & 0 & 0.2 & 32 / 7 \\
ogbn-products & 2,449,029 & 61,859,140 & 100   & 47 & 3$\times$256 BN & 3$\times$256 BN & 0.003 & 0 & 0.5 & 64 / 7 \\
\bottomrule
\end{tabular}
\end{table*}

\noindent\textbf{Teacher-free implementation.}
For $\lambda=0$, we skip the distillation pass entirely rather than executing a zero-weighted
soft-label loss, because the latter would still apply optimizer updates through weight decay and
momentum in the inherited two-pass training loop. All teacher-free results use this corrected path.

\begin{table}[!t]
\centering
\footnotesize
\caption{Graph-classification datasets and student hyperparameters (all use rank $r{=}64$).
TU splits use 10-fold CV, \textsc{ogbg-molhiv} the official scaffold split.}
\label{tab:hyper_graph}
\setlength{\tabcolsep}{4pt}
\begin{tabular}{lrrcccc}
\toprule
Dataset & graphs & avg.\ nodes & hidden & layers & batch & epochs \\
\midrule
\textsc{ogbg-molhiv} & 41,127 & 25.5  & 300 & 3 & 256 & 100 \\
PROTEINS             & 1,113  & 39.1  & 128 & 2 & 128 & 100 \\
DD                    & 1,178  & 284.3 & 128 & 2 & 64  & 100 \\
IMDB-B                & 1,000  & 19.8  & 128 & 2 & 128 & 100 \\
REDDIT-B              & 2,000  & 429.6 & 128 & 2 & 64  & 100 \\
NCI1                  & 4,110  & 29.9  & 128 & 2 & 128 & 100 \\
\bottomrule
\end{tabular}
\vspace{-1em}
\end{table}

\subsection{Proof of Proposition~\ref{prop:equiv} (symmetry inheritance)}
\label{app:proof_equiv}
\begin{proof}
Under relabeling, node features and equivariant fingerprints are permuted consistently:
$\vx_{\pi(v)}(\pi\gG)=\vx_v(\gG)$ and $\vz_{\pi(v)}(\pi\gG)\allowbreak=\vz_v(\gG)$. Because the generator
and predictor are shared and applied node-wise, corresponding nodes receive identical inputs and
produce identical outputs, proving equivariance. For graph-level conditioning, an invariant
fingerprint produces identical effective weights after relabeling; the shared node-wise encoder is
equivariant and mean pooling is invariant, proving graph-level invariance.
\end{proof}

\subsection{Proof of Proposition~\ref{prop:film} (weight generation escapes row-rescaling)}
\label{app:proof_film}
\begin{proof}
Each row of $\mD\mW_0$ is a scalar multiple of the corresponding row of $\mW_0$, so
$\mathcal{T}_{\mathrm{diag}}(\mW_0)$ contains only row-wise rescalings. By assumption, some basis
element $\mathbf{u}_i\mathbf{v}_i^\top$ lies outside this set. Choosing $\va=\mathbf{e}_i$ yields
that perturbation, so $\mathcal{T}_{\method}(\mU,\mV)$ is not contained in
$\mathcal{T}_{\mathrm{diag}}(\mW_0)$. Therefore the corresponding effective weight cannot be
represented as $\operatorname{diag}(\gamma)\mW_0$.
\end{proof}

\subsection{Proof of Corollary~\ref{cor:mapping}}
\label{app:proof_cor}
\begin{proof}
Fix the shared basis from Eq.~\eqref{eq:shared_basis_assumption} and define the linear operator
$\mathcal B:\mathbb R^r\to\mathbb R^{o\times d}$ by
$\mathcal B(\mathbf c)=\mU\operatorname{diag}(\mathbf c)\mV^\top$.
Let $C_{\mathcal B}=\sup_{\|\mathbf c\|_2=1}\|\mathcal B(\mathbf c)\|_F$, which is finite because the
spaces are finite-dimensional. The coefficient map $\mathbf a$ is continuous on the compact domain
$\mathcal Z$. By universal approximation, for any $\eta>0$ there exists a two-layer MLP
$\vs:\mathcal Z\to\mathbb R^r$ satisfying
\[
\sup_{\vz\in\mathcal Z}\|\vs(\vz)-\mathbf a(\vz)\|_2
\leq \frac{\eta}{C_{\mathcal B}+1}.
\]
Consequently,
\[
\sup_{\vz\in\mathcal Z}
\|\mathcal B(\vs(\vz))-\mathcal B(\mathbf a(\vz))\|_F
\leq C_{\mathcal B}\frac{\eta}{C_{\mathcal B}+1}
\leq \eta.
\]
Combining this inequality with Eq.~\eqref{eq:shared_basis_assumption} and the triangle inequality gives
\[
\sup_{\vz\in\mathcal Z}
\left\|\mW^{*}(\vz)-
\left(\mW_0+\mU\operatorname{diag}(\vs(\vz))\mV^\top\right)\right\|_F
\leq \delta_r+\eta.
\]
Finally, the uniform $L$-Lipschitz property of each node loss converts this weight error into
\[
\ell_v\!\left(\mW_0+\mU\operatorname{diag}(\vs(\vz_v))\mV^\top\right)
-\ell_v\!\left(\mW^{*}(\vz_v)\right)
\leq L(\delta_r+\eta),
\]
which proves Cor.~\ref{cor:mapping} for every node $v$.
\end{proof}

\subsection{Implementation Details and Hyperparameters}
\label{app:hyper}

All models are trained with Adam for up to $200$ epochs with early stopping on validation accuracy
(patience $50$); teachers use neighbor sampling: fan-out $5{,}10{,}15$ on the OGB graphs
and $5{,}5$ on the CPF benchmarks. Architectures and per-dataset hyperparameters
follow the released GLNN configurations and are listed in Table~\ref{tab:hyper}; the ranks are
\method's, with the Low-rank control always rank-matched. The +LapPE arm conditions on the $32$
leading eigenvectors of the normalized Laplacian; SGC selects $K\in\{2,3,5,10\}$ on validation
per seed with learning rate swept over $\{0.01,\allowbreak 0.05,\allowbreak 0.1,\allowbreak
0.2\}$ and a $1000$-epoch budget
(patience $300$), keeping each dataset's best configuration; the DeepWalk arms use each
dataset's NOSMOG walk settings.
We write $n_e$ for the number of random seeds; the main text uses $n_e{=}10$ and appendix ablations $n_e{=}3$ unless a table states otherwise, and ablation rows are matched within each table.
\subsection{Inference Latency Benchmark}
\label{app:latency}

Figure~\ref{fig:efficiency} measures wall-clock latency on a single RTX 3090, GPU-synchronized,
after a five-call warm-up and averaged over $20$ timed calls per configuration, at a fixed node
count ($N=169{,}343$, matching \textsc{ogbn-arxiv}) with edge count swept over eight densities. All
three models, the SAGE teacher, GLNN, and \method, are passed through the same \texttt{torch.compile}
call before timing; the teacher's message passing, implemented with DGL's custom sparse kernels,
hits an uncompilable operation (\texttt{torch.\_C.\_to\_dlpack}) and falls back to eager execution
for that part, so it sees little benefit, while both message-passing-free students compile cleanly.

\method's forward pass evaluates the stacked-projection realization described in \S\ref{sec:swg}:
$\mW_0\vx_v$ and $\mV^\top\vx_v$ are computed as one matrix-vector product against the concatenated
matrix $[\mW_0;\mV^\top]$, followed by the element-wise scaling and the $\mU(\cdot)$ projection. At
\textsc{ogbn-arxiv}'s scale and edge density, this exact realization alone (still eager) reduces
per-layer latency from $8.23$\,ms (three separate projections) to $6.69$\,ms ($19\%$ faster);
compiling on top of the stacked realization reaches $5.2$\,ms ($37\%$ faster than the uncompiled,
unstacked baseline), the configuration plotted in Figure~\ref{fig:efficiency}.

\subsection{System-Level Context on the CPF Benchmarks}
\label{app:cpf_systems}

Table~\ref{tab:cpf_systems} completes the system-level view of \S\ref{sec:exp_scope} on the five
CPF datasets, with the same signal pairing as Table~\ref{tab:ogb_systems}. The \method\ rows are
tuned per dataset on validation accuracy, mirroring the external systems' own tuning, and their
ho-obs fingerprints are built from observed edges only, with the random-walk encoding computed on
the observed graph augmented with the held-out node and the DeepWalk signal extended through
observed neighbors. The published systems stay ahead at this scale through their additional
distillation objectives, while positional input is the decisive ingredient on \textsc{Cora},
where adding the DeepWalk input lifts \method's held-out accuracy by nearly nine points. The
mechanism this paper isolates separates from the shared student only at OGB scale, which
Table~\ref{tab:ogb_systems} reports.

\begin{table}[t]
\centering
\footnotesize

\caption{System-level context on the five CPF benchmarks, transductive (top) and ho-obs (bottom),
paired as in Table~\ref{tab:ogb_systems}. External numbers as reported by
TINED~\cite{zhou2025tined}; \method\ rows are per-dataset tuned.}

\label{tab:cpf_systems}
\setlength{\tabcolsep}{2pt}
\resizebox{\columnwidth}{!}{%
\begin{tabular}{@{}lccccc@{}}
\toprule
Method & Cora & Citeseer & Pubmed & A-computer & A-photo \\
\midrule
\multicolumn{6}{@{}l}{\emph{Transductive}} \\
TINED~\cite{zhou2025tined} & 82.63$\pm$1.57 & 74.43$\pm$1.53 & 77.09$\pm$2.14 & 85.18$\pm$1.12 & 93.97$\pm$0.53 \\
\method & 80.60$\pm$2.16 & 72.11$\pm$1.46 & 76.25$\pm$1.88 & 82.42$\pm$1.89 & 92.06$\pm$0.93 \\
\addlinespace[2pt]
NOSMOG~\cite{tian2023nosmog} & 83.04$\pm$1.26 & 73.78$\pm$1.54 & 77.34$\pm$2.36 & 84.04$\pm$1.01 & 93.36$\pm$0.69 \\
\method + DeepWalk & 80.57$\pm$2.45 & 72.18$\pm$1.32 & 76.25$\pm$1.96 & 82.60$\pm$2.02 & 91.23$\pm$0.94 \\
\addlinespace[2pt]
TINED+~\cite{zhou2025tined} & 83.70$\pm$1.02 & 75.39$\pm$1.59 & 77.75$\pm$3.14 & 84.82$\pm$1.58 & 94.05$\pm$0.39 \\
\method + DW input & 81.08$\pm$1.60 & 70.16$\pm$1.75 & 75.78$\pm$2.25 & 82.63$\pm$1.67 & 90.95$\pm$1.20 \\
\midrule
\multicolumn{6}{@{}l}{\emph{Held-out nodes (ho-obs)}} \\
TINED~\cite{zhou2025tined} & 74.38$\pm$1.28 & 72.68$\pm$1.97 & 75.64$\pm$3.02 & 82.83$\pm$1.45 & 91.96$\pm$0.72 \\
\method & 71.01$\pm$3.11 & 70.17$\pm$2.27 & 74.94$\pm$2.72 & 80.59$\pm$1.32 & 90.09$\pm$1.14 \\
\addlinespace[2pt]
NOSMOG~\cite{tian2023nosmog} & 81.36$\pm$1.53 & 70.30$\pm$2.30 & 75.87$\pm$3.32 & 84.36$\pm$1.57 & 92.61$\pm$1.09 \\
\method + DeepWalk & 71.22$\pm$3.15 & 69.34$\pm$2.87 & 75.53$\pm$2.49 & 83.06$\pm$1.74 & 91.19$\pm$0.78 \\
\addlinespace[2pt]
TINED+~\cite{zhou2025tined} & 81.50$\pm$2.54 & 74.20$\pm$1.67 & 76.30$\pm$2.95 & 85.45$\pm$1.60 & 93.27$\pm$0.85 \\
\method + DW input & 79.74$\pm$2.66 & 70.36$\pm$2.50 & 73.89$\pm$2.86 & 83.44$\pm$1.56 & 91.43$\pm$1.58 \\
\bottomrule

\end{tabular}%
}
\vspace{-0.8em}
\end{table}

\subsection{Mechanism Ablations}
\label{app:capacity}

\noindent\textbf{Generator capacity.}
Fig.~\ref{fig:abrank} plots the full conditioning-rank sweep. The generator-depth sweep (depth
$1$--$3$, with and without LayerNorm) confirms that depth-$1$ already captures the gain and added
depth is inert. Table~\ref{tab:app_B_lora} sweeps rank on \textsc{ogbn-products}.

\noindent\textbf{Mounting depth.} \method\ keeps the output head unadapted, following the LoRA
convention~\cite{hu2022lora}; the Mapping Networks construction~\cite{sen2026mapping} instead
generates every layer of the target network, biases included. An \textsc{ogbn-arxiv} ablation
($r32$, $n_e{=}3$) interpolates between the two: generating only the first layer, both hidden
layers (the reported configuration), and additionally the output head gives $68.7$, $69.1$, and
$69.6$ transductively, with the deepest mounting also best under ho-full and preferred by validation in both settings.
We keep the non-output \method{} convention throughout for comparability across tables.

\begin{table}[t]
\centering
\footnotesize
\caption{Rank behavior on the OGBN graphs (\%). (a) \method\ rank sweep on \textsc{ogbn-products}.
(b) The unconditioned Low-rank control never improves on GLNN, and collapses at $r8$.}
\label{tab:app_B_lora}
\begin{tabular}{lcc}
\toprule
Variant & tran & ho-full \\
\midrule
\multicolumn{3}{@{}l}{(a) \emph{conditioned: rank sweep on \textsc{ogbn-products}}} \\
\method\ $r32$                        & 70.10 & 69.64 \\
\textbf{\method\ $r64$}               & \textbf{70.63} & \textbf{70.06} \\
\midrule
\multicolumn{3}{@{}l}{(b) \emph{unconditioned: Low-rank control vs.\ GLNN, both graphs}} \\
Low-rank $r8$ (arxiv)              & 53.60 & 53.45 \\
Low-rank $r32$, matched (arxiv)    & 60.08 & 58.05 \\
GLNN (arxiv)                       & 64.81 & 58.98 \\
\addlinespace
Low-rank $r8$ (products)           & 52.57 & 52.52 \\
Low-rank $r64$, matched (products) & 63.90 & 63.41 \\
GLNN (products)                    & 64.60 & 64.16 \\
\bottomrule
\end{tabular}
\vspace{-1em}
\end{table}

\noindent\textbf{Learned design components.}
\label{app:design}
The basis must be learned jointly with the generator; it cannot be inherited from
the teacher's weights. Equation~\eqref{eq:structlora} leaves $\mW_0$, $\mU$, $\mV$ free, while two natural families of
alternatives either fix them from the teacher or make them outputs of the generator. Table~\ref{tab:design}
follows a constraint spine on \textsc{ogbn-arxiv} ($r32$, $n_e{=}3$).
It first initializes $\mW_0$ from the teacher's corresponding layer, in the spirit of teacher
injection~\cite{zhou2025tined}, then additionally freezes it, ties $\mU,\mV$ to the frozen
base's top singular vectors, and finally freezes those too. This yields a fully constrained form in which the generator can only reweight the teacher's singular spectrum, the SVD-style adapter parameterisation that AdaLoRA uses to allocate rank budget across layers~\cite{zhang2023adalora}: $\mW_{\mathrm{eff}}=\mU_w\,\mathrm{diag}(\bm\sigma+\vs(\vz))\,\mV_w^\top$. The spine degrades from
top to bottom, monotonically apart from one within-noise tie (Table~\ref{tab:design}). The controls sharpen the reading: it is \emph{freezing}, not teacher initialization, that hurts. This contrasts with LoRA, where the two factors play asymmetric roles, and the down-projection can be left at its random initialization at little cost~\cite{zhu2024asymmetry}. Orthogonality constraints also cost transductively, and the random fully frozen variant falls to the frozen-map floor of \S\ref{sec:exp_ablation}.

A second alternative makes the generator emit the factors themselves,
$\Delta(v)=\mA(\vz_v)\,\mathbf{B}(\vz_v)$: the direct per-node adaptation of standard
factor-emitting conditional hypernetworks~\cite{ha2017hypernetworks,debrabandere2016dynamic},
serving as an external hypernetwork baseline for this deployment regime.
This variant does not improve over the structure-free baseline. Both the shared basis and the generated coefficients therefore appear to matter, though this arm alone does not separate their contributions.

\begin{table}[t]
\centering
\footnotesize
\caption{Design-constraint spine on \textsc{ogbn-arxiv} ($r32$, \%).}
\label{tab:design}
\begin{tabular}{lcc}
\toprule
Constraint added (cumulative) & tran & ho-full \\
\midrule
none (\method, all learned) & \textbf{69.00} & \textbf{64.50} \\
$\mW_0\!\leftarrow$ teacher & 68.85 & 64.01 \\
\;+ $\mW_0$ frozen & 65.21 & 63.35 \\
\;+ $\mU,\mV\!\leftarrow$ SVD($\mW_0$) & 65.54 & 63.45 \\
\;+ $\mU,\mV$ frozen (fully constrained) & 61.87 & 60.99 \\
\midrule
\emph{control:} SVD basis, orthogonalized & 64.72 & 63.71 \\
\emph{control:} teacher init, nothing frozen & 68.89 & 64.07 \\
\emph{control:} random $\mW_0,\mU,\mV$, all frozen & 59.54 & 58.54 \\
\bottomrule
\end{tabular}
\vspace{-1em}
\end{table}

\subsection{Weight-Manifold / FIM Geometry}
\label{app:fim}

\begin{figure}[h]
\centering
\includegraphics[width=0.58\linewidth]{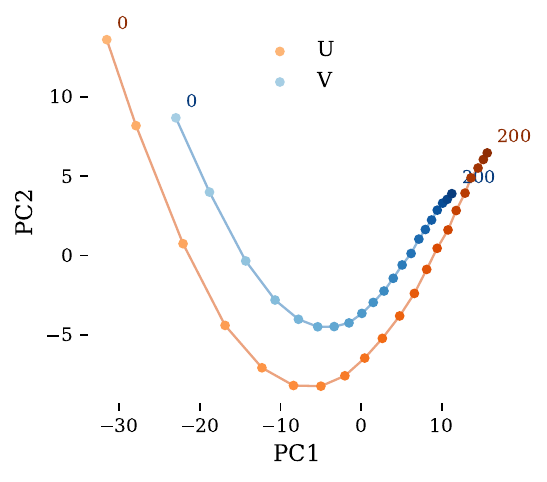}
\caption{PCA trajectories of $\mathrm{vec}(\mU_t)$ and $\mathrm{vec}(\mV_t)$ over training on
\textsc{ogbn-arxiv} ($21$ snapshots, every $10$ epochs; each factor on its own top-2 basis, color
light to dark).}
\label{fig:uv_manifold}
\end{figure}

This appendix provides the protocols and full results behind \S\ref{sec:exp_geometry}. All tests use the
empirical Fisher information matrix (FIM) $F=\mathbb{E}_v[g_v g_v^{\top}]$ with
$g_v=\partial\mathrm{KL}/\partial W_0$ ($N{=}1200$ nodes). \textbf{(1) Subspace overlap}
(Table~\ref{tab:app_F_overlap}; trained \method\ with a \emph{learned} low-rank basis $U,V$):
we report the overlap between the structure-reachable subspace and the empirical FIM's
top-$\kappa$ eigenspace relative to a random subspace ($1.0$ = random; the random baseline
$\mathrm{overlap}=\kappa/d$ was validated exactly). As rank grows, the learned
subspace develops substantially lower-than-random overlap with the leading
empirical-Fisher eigenspace (at $r64$ it is $\approx0.2\times$ random). Since empirical Fisher is
not identical to the loss Hessian, this is a statement about eigenspace overlap under our
diagnostic, not a direct claim about global flatness of the loss. \textbf{(2) Functional per-node test}
(Table~\ref{tab:func_main}, \S\ref{sec:exp_geometry}; ECDFs in Fig.~\ref{fig:func_ecdf}; single
run, $N{=}8000$ nodes, five random permutations). In all configs, the directional cosine of
$\Delta W(v)$ against $-g_v$ is positive and node-specific (own $\gg$ swap $\approx0$), though
small ($\sim0.003$); a single raw cosine in a $\sim\!10^4$-dimensional weight space is dominated by
near-orthogonal noise directions regardless of effect size, so we treat the sign and the
own-vs-swap gap, not the magnitude, as the diagnostic. The example node in
Fig.~\ref{fig:weightspace_schematic}b is illustrative rather than a population estimate: it was
selected by a pre-registered rule, the $75$th percentile of the mechanism-off $\Delta$KL among a
$2{,}000$-node sample, without inspecting its surface or trajectory; population-level statistics
are the subject of this test. Complementing the overlap curves of
Fig.~\ref{fig:weightspace_schematic}c, Fig.~\ref{fig:uv_manifold} shows the factors themselves:
the PCA trajectories of $\mathrm{vec}(\mU_t)$ and $\mathrm{vec}(\mV_t)$ across the same $21$
snapshots trace smooth low-dimensional curves, the factor-level analogue of the per-node code
evolution in Fig.~\ref{fig:manifold}.

\begin{table}[t]
\centering
\footnotesize
\caption{Subspace-overlap test on \textsc{ogbn-arxiv} (transductive). Overlap between the
structure-reachable subspace and the FIM top-$\kappa$ eigenspace, normalized so $1.0=$ random.
The $r32$ thick row is exploratory.}
\label{tab:app_F_overlap}
\begin{tabular}{lccccc}
\toprule
Config & $\kappa{=}8$ & 16 & 32 & 64 & 128 \\
\midrule
$r16$        & 1.41 & 1.10 & 1.48 & 1.15 & 1.09 \\
$r32$        & 0.33 & 0.52 & 0.51 & 0.45 & 0.46 \\
$r64$        & 0.23 & 0.24 & 0.21 & 0.21 & 0.21 \\
$r32$ thick  & 0.67 & 0.70 & 0.60 & 0.52 & 0.56 \\
\bottomrule
\end{tabular}
\vspace{-1em}
\end{table}

\textbf{(3) Restricted Fisher, fixed orthonormal basis.} Tests (1) and (2) address different
aspects; a third test asks a further question. Overlap test (1) asks whether the low-rank subspace
\emph{as a whole} has above- or below-random overlap with the leading empirical-Fisher
eigenspace. Under a variant of \method\ that fixes $U,V$ to a random orthonormal frame at initialization (only the base weight and the generator $s(\cdot)$ train), that subspace is a frozen random draw independent of training. Asking whether
\emph{it} aligns with the global FIM top-$\kappa$ eigenspace would therefore answer a
question about the random draw, not about what the generator learned. We
instead ask a different, non-tautological question restricted to the fixed
door itself: writing $F_S=S^{\top}FS$ for the FIM restricted to the subspace
$S$ spanned by the frozen basis, does the generator's cross-node output
variance concentrate on $F_S$'s \emph{own} most-informative directions? Across
three independently drawn random bases and two structural fingerprints (a
local one and a spectral one) on \textsc{ogbn-arxiv}, the population-level
correlation between a direction's informativeness under $F_S$ and the
generator's output variance along it is consistently positive and significant
(Spearman $\rho=+0.48$ to $+0.80$, permutation $p{<}0.01$ in five of six runs).
On average, $62$--$84\%$ of each node's generated-update energy falls in the
upper half of $F_S$'s eigenspectrum. Thus, while the learned low-rank subspace as
a whole has substantially lower-than-random overlap with the leading empirical-Fisher eigenspace
at larger ranks (test 1), the signal placed \emph{within} a subspace, even an arbitrary fixed one,
is not distributed arbitrarily. It preferentially routes through whichever directions
are locally informative given the basis it was handed.

\begin{figure}[t]
\centering
\includegraphics[width=0.68\linewidth]{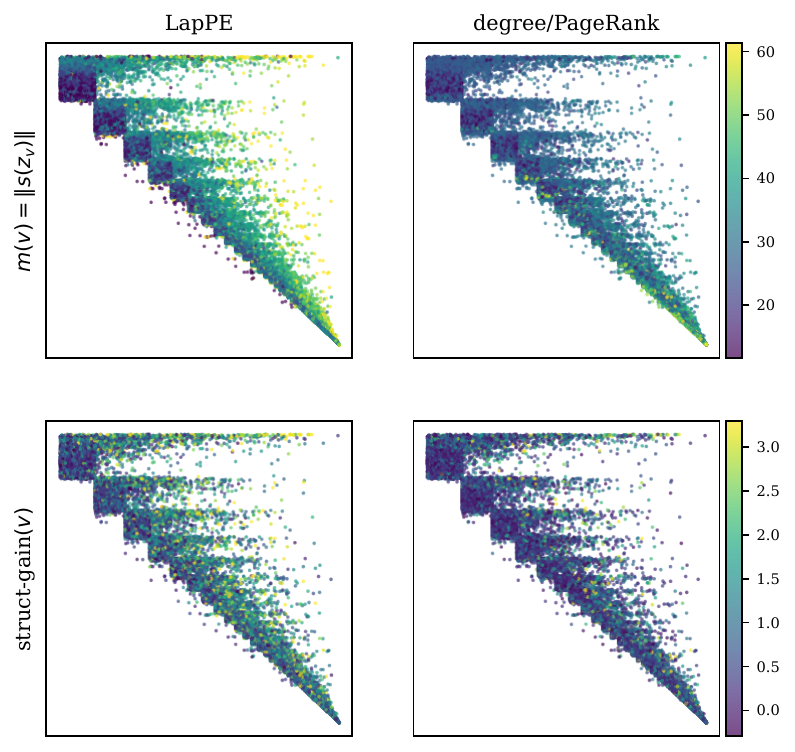}
\caption{Illustrative, not evidentiary: nodes at a fixed, graph-only Laplacian-eigenvector layout
(identical in all panels), colored by the basis-invariant $m(v)$ (top) and struct-gain$(v)$
(bottom), under LapPE (left) and degree/PageRank (right).}
\label{fig:manifold_invariant}
\end{figure}

The three tests reveal one mechanism from complementary angles. Structure-conditioning
is a \emph{node-specific} weight nudge that, as a whole, has low overlap with the leading
empirical-Fisher eigenspace of the \emph{average} loss. This is consistent with why
teacher soft labels may already cover it (\S\ref{sec:exp_teacher}). At the same time, it remains
organized with respect to sensitivity directions within the restricted empirical-Fisher subspace,
and is functionally strong per node. This is also consistent with the functional $\Delta\mathrm{KL}$, but requires
$z_v$ at inference. Figure~\ref{fig:manifold_invariant} visualizes $m(v)$ and struct-gain$(v)$
across the graph for one random basis.

\subsection{Full Graph Classification (ogbg-molhiv)}
\label{app:graphcls}

\begin{figure}[tb]
\centering
\includegraphics[width=0.62\linewidth]{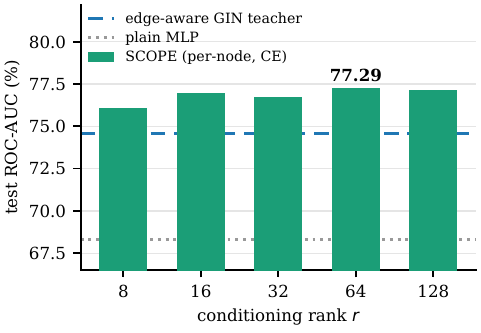}
\caption{ROC-AUC vs.\ rank for the per-node \method\ student on \textsc{ogbg-molhiv}, against the GIN
teacher and plain-MLP references.}
\label{fig:graphcls_rank}
\vspace{-1em}
\end{figure}

This appendix details the per-graph spectral comparison point for the primary per-node
instantiation of \S\ref{sec:exp_graph}. The construction follows \S\ref{sec:graph_level}, with a
$16$-dimensional spectral block in $z_{\mathrm{graph}}$; all numbers are ROC-AUC. The main
comparison (Table~\ref{tab:app_G}a) shows the per-graph \method\ (CE) student clearly beats the
plain MLP. Unlike in node classification, the teacher is redundant here too: KD never
significantly helps, the largest KD$-$CE test difference being well within seed noise.

\begin{table}[t]
\centering
\footnotesize

\caption{Per-graph spectral \method\ on \textsc{ogbg-molhiv} (ROC-AUC); student configuration as in
Table~\ref{tab:hyper_graph}. The $r64$ (CE) and edge-aware GIN rows are those reported in
Table~\ref{tab:graph}; (b) is a separate sweep.}

\label{tab:app_G}
\begin{tabular}{lcc}
\toprule
 & val & test \\
\midrule
\multicolumn{3}{@{}l}{(a) \emph{main comparison}} \\
GIN teacher, plain (ref) & 80.56 & 73.57 \\
plain MLP (CE)           & 72.18 & 68.64 \\
plain MLP (KD)           & 72.07 & 66.89 \\
per-graph \method\ (CE, $r64$) & 79.67 & \textbf{71.58} \\
per-graph \method\ (KD)  & 79.45 & 70.90 \\
\midrule
\multicolumn{3}{@{}l}{(b) \emph{rank sweep, per-graph \method\ (CE)}} \\
plain MLP (CE)   & 72.15 & 68.31 \\
$r8$             & 79.43 & 72.45 \\
$r16$            & 78.91 & \textbf{73.04} \\
$r32$            & 81.11 & 71.59 \\
$r64$            & 80.69 & 71.98 \\
$r128$           & 79.70 & 71.13 \\
\midrule
\multicolumn{3}{@{}l}{(c) \emph{strong-teacher control}} \\
strong teacher, edge-aware          & 80.55 & 74.61 \\
plain MLP (KD, strong T)           & 75.20 & 65.29 \\
per-graph \method\ (CE, $r64$)      & 79.67 & 71.58 \\
per-graph \method\ (KD, strong T)  & 79.49 & 70.62 \\
\midrule
\multicolumn{3}{@{}l}{(d) \emph{$z$-granularity ($r32$, CE)}} \\
per-node (local)                & 80.22 & \textbf{76.72} \\
per-graph (spectral broadcast)  & 80.23 & 72.31 \\
concat [$z_{\mathrm{node}}\|z_{\mathrm{graph}}$] & 82.28 & 72.71 \\
\bottomrule
\end{tabular}
\end{table}

A rank sweep (Table~\ref{tab:app_G}b) strengthens the test gain: every rank beats
the plain MLP by $+3$ to $+4.7$ test points, non-monotonically, with a sweet spot at $r16$; the
main-text comparison (Table~\ref{tab:graph}) uses an independently rerun $r64$ point, within half
a point of this sweep's $r64$ row. A strong-teacher control (Table~\ref{tab:app_G}c, edge-aware
GIN with BondEncoder) shows CE $\ge$ KD persists with a genuinely strong teacher, and distilling
it even \emph{hurts} the plain MLP, the failure mode anticipated in \S\ref{sec:exp_graph}.

Finally, granularity matters (Table~\ref{tab:app_G}d): a \emph{per-node} local fingerprint beats the
per-graph broadcast ($+4.4$ test) and even beats concatenating both fingerprints together. A separate per-node rank sweep (Fig.~\ref{fig:graphcls_rank}) reaches its best reported point at
$r64$, test $77.29\%$, exceeding the strong edge-aware GIN reference ($74.61\%$) by $2.7$ points in
this run, without a teacher and online message passing.

\clearpage
\end{document}